\documentclass[journal]{IEEEtran}

\usepackage{amsmath}
\usepackage{totpages}
\usepackage{graphicx}
\usepackage{xcolor}
\usepackage{subcaption}
\usepackage{url}
\usepackage{float}
\usepackage{algpseudocode}
\usepackage{algorithm}
\usepackage{multirow}
\usepackage{tikz}
\usepackage{wrapfig}
\usepackage{hyperref}
\usepackage{cleveref}
\usepackage{makecell}
\usepackage{xspace}
\usepackage[normalem]{ulem}
\usepackage{enumitem}
\usepackage{pifont}
\usepackage{caption}
\usepackage{makecell}
\usepackage[utf8]{inputenc}
\usepackage{amssymb}
\usepackage{svg}
\usepackage{booktabs}
\usepackage{siunitx}
\usepackage{textcomp}
\usepackage{cite}

\begin{document}

\title{Mon3tr: Monocular 3D Telepresence with Pre-built Gaussian Avatars as Amortization}

\author{Fangyu~Lin, 
        Yingdong~Hu, 
        Zhening~Liu, 
        Yufan~Zhuang, \\
        Zehong~Lin,~\IEEEmembership{Member,~IEEE,}
        and~Jun~Zhang,~\IEEEmembership{Fellow,~IEEE}
\thanks{The authors are with the Department of Electronic and Computer Engineering, The
Hong Kong University of Science and Technology, Hong Kong (e-mails: flinao@connect.ust.hk; yhudj@connect.ust.hk; zhening.liu@connect.ust.hk; yzhuangah@connect.ust.hk; eezhlin@ust.hk; eejzhang@ust.hk). \textit{(Corresponding authors: Zehong Lin; Jun Zhang.)}}
\thanks{}
\thanks{}}

\markboth{Journal of \LaTeX\ Class Files
}%
{Shell \MakeLowercase{\textit{et al.}}: Mon3tr: Monocular 3D Telepresence with Pre-built Gaussian Avatars as Amortization}

\maketitle
\begin{abstract}
Immersive telepresence aims to transform human interaction in AR/VR applications by enabling lifelike full-body holographic representations for enhanced remote collaboration. However, existing systems rely on hardware-intensive multi-camera setups and demand high bandwidth for volumetric streaming, limiting their real-time performance on mobile devices. To overcome these challenges, we propose \textbf{Mon3tr}, a novel \textbf{\uline{Mon}}ocular \textbf{\uline{3}}D \textbf{\uline{t}}elep\textbf{\uline{r}}esence framework that integrates 3D Gaussian splatting (3DGS) based parametric human modeling into telepresence for the first time. Mon3tr adopts an amortized computation strategy, dividing the process into a \textit{one-time offline multi-view reconstruction phase} to build a user-specific avatar and \textit{a monocular online inference phase} during live telepresence sessions. A single monocular RGB camera is used to capture body motions and facial expressions in real time to drive the 3DGS-based parametric human model, significantly reducing system complexity and cost. The extracted motion and appearance features are transmitted at $<$ 0.2 Mbps over WebRTC’s data channel, allowing robust adaptation to network fluctuations. On the receiver side, e.g., Meta Quest 3, we develop a lightweight 3DGS attribute deformation network to dynamically generate corrective 3DGS attribute adjustments on the pre-built avatar, synthesizing photorealistic motion and appearance at $\thicksim$ 60 FPS. Extensive experiments demonstrate the state-of-the-art performance of our method, achieving a PSNR of $>$ 28 dB for novel poses, an end-to-end latency of $\thicksim$ 80 ms, and $>$ 1000$\times$ bandwidth reduction compared to point-cloud streaming, while supporting real-time operation from monocular inputs across diverse scenarios. Our demos can be found at \url{https://mon3tr3d.github.io}.
\end{abstract}


\IEEEpeerreviewmaketitle

\section{Introduction}
Real-time immersive telepresence is poised to revolutionize digital human interaction and serves as a foundation for next-generation augmented and virtual reality (AR/VR) applications. By enabling lifelike full-body holographic avatars, telepresence systems facilitate authentic social presence and enhance remote collaboration \cite{6g1, 6g2}. Unlike traditional 2D video conferencing, which limits users to fixed viewpoints, 3D telepresence allows users to perceive and interact with remote participants in a six degrees of freedom (6-DoF) framework. This rich spatial interaction preserves subtle yet crucial non-verbal cues, such as gestures, posture, and facial expressions, enabling transformative applications ranging from immersive professional training and telesurgery to virtual conferencing and social connectivity in the metaverse \cite{volumetric}.

A typical 3D telepresence pipeline comprises three key components: a transmitter, a data channel, and a receiver. The transmitter is equipped with multiple RGB/RGB-D cameras and GPU-equipped processors to capture and reconstruct the 3D geometry and appearance of a human subject. The data channel provides a well-configured communication link to transmit 3D information in real time \cite{holoportation, metastream, project_straline, farfetchfusion, virtualcube}. The receiver, typically a VR/AR headset or an autostereoscopic display, renders the volumetric content for immersive viewing. In interactive scenarios, the roles of the transmitter and receiver can dynamically switch between participants, enabling bidirectional communication. Traditional systems mainly rely on classic 3D representations such as dense meshes or point clouds to model human geometry. However, these representations require precise and dense reconstruction, which is computationally intensive and sensitive to sampling noise, resulting in limited visual quality and robustness for daily use.

In recent years, researchers have explored advanced scene representations. Neural radiance fields (NeRFs) \cite{nerf} represent a breakthrough by implicitly modeling scenes as continuous volumetric functions parameterized by coordinate-based neural networks. Representing humans as posed NeRFs can significantly improve rendering quality \cite{humannerf}. Nevertheless, the reliance of NeRFs on multi-layer perceptrons (MLPs) introduces a spectral bias toward low-frequency details and incurs high inference latency. This latency is prohibitive for interactive telepresence applications, which require end-to-end delays below 100 ms to maintain natural interaction and ensure a seamless user experience \cite{marvel}. More recently, 3D Gaussian splatting (3DGS) \cite{3dgs} has emerged as a promising alternative. Its efficient point-based rasterization allows for both high-fidelity rendering and real-time performance. However, building telepresence systems based on 3DGS faces a critical bottleneck: the large data size per frame leads to prohibitive communication overhead, making it challenging to support real-time operation over conventional networks.

\begin{figure*}[t!]
    \centering
    \includegraphics[width=\textwidth]{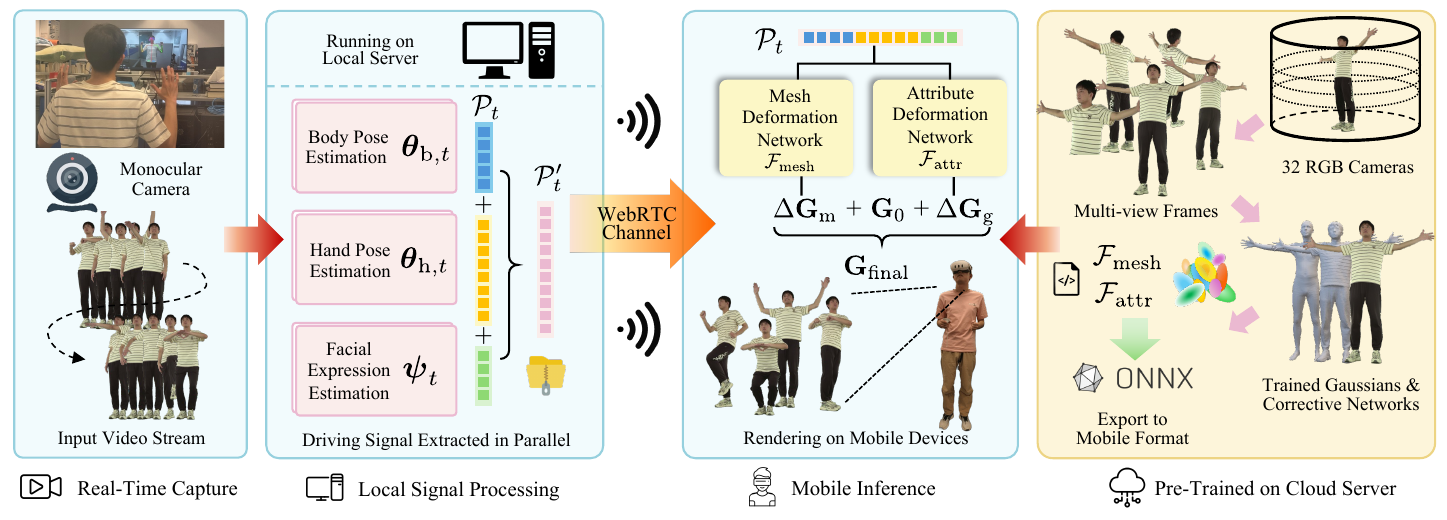}
    \vspace{-21pt}    \caption{System overview of \textbf{Mon3tr}. Before the online immersive telepresence or conferencing, Mon3tr pre-builds a photorealistic animatable avatar based on pre-recorded video clips, which is then uploaded to a cloud server for subsequent use. During runtime, a monocular RGB camera captures and estimates body pose, hand gesture and facial expression parameters \cite{gvhmr, hamer, smirk} in parallel and sends them to the VR viewer via a Wi-Fi router with controlled bandwidth \cite{tc}. The pre-built avatar is then driven in real time at $\thicksim$ 60 FPS on the device.}
    \label{fig:pipeline}
\end{figure*}

To address the communication bottleneck, several approaches have been proposed. Some methods focus on compression and encoding schemes \cite{l3gs, dass, ffd3dgsc, mega} to reduce the size of transmitted data. Nevertheless, they struggle with the inherently large per-frame data volumes of 3D representations, limiting their effectiveness in bandwidth-constrained mobile networks. Alternative methods attempt to simplify the representation itself, such as reducing 3DGS to 2DGS with fewer attributes \cite{2dgs}. While these methods lower bandwidth requirements, they often fail to consistently maintain a high quality of experience (QoE), especially under fluctuating network conditions \cite{vivo}. Another line of works, exemplified by MagicStream \cite{magicstream}, propose transmitting compact semantic information about the human body geometry and lighting conditions rather than raw 3D data. These semantic cues enable neural rendering on the receiver side, drastically reducing bandwidth requirements to as low as 0.2 Mbps. Although promising, this approach depends on costly multi-camera RGB-D capture setups, which require precise calibration and complex synchronization across multiple sensors, and typically cannot achieve photorealistic rendering fidelity.

To overcome these limitations, we introduce \textbf{Mon3tr}, a novel monocular 3D telepresence framework that breaks the traditional trade-off between visual quality, real-time rendering, and communication overhead. The key insight behind \textbf{Mon3tr} is amortized computation: instead of performing costly full-scene reconstruction and transmission for every frame, we conduct a one-time offline process to build a high-fidelity, personalized, animatable 3DGS avatar rich in geometric and appearance details. During live telepresence sessions, only a compact set of animation and expression parameters is transmitted to drive the pre-built avatar in real time. This approach effectively amortizes the significant cost of generating photorealistic avatars over the entire session, decoupling visual quality from per-frame computational and network load. As a result, Mon3tr enables cinematic realism on consumer hardware such as personal computers (PCs) and Meta Quest 3 \cite{quest3}, achieving a real-time streaming rate at $\thicksim$ 60 FPS while drastically reducing bandwidth requirements. 

Our work makes the following key contributions to enable efficient amortized computation for high-fidelity monocular 3D telepresence:
\begin{itemize}
\item \textbf{A High-fidelity Human Mesh Template:}
We propose a new parametric human model, SPMM3 (skinned person model for Mon3tr), to obtain a personalized animatable mesh template that captures both coarse and fine-grained details on the human body. For body modeling, we reconstruct a detailed human body mesh that can represent more details of clothes and hair from multi-view images. By combining the captured body mesh with dedicated hand and face models, the expressiveness of the human model is further extended with vivid expressions and flexible gestures.

\item \textbf{An Animatable 3DGS-based Avatar Representation:} 
We build a 3DGS-based avatar by binding Gaussians to the proposed mesh template SPMM3. To capture realistic non-rigid deformations, we introduce a lightweight mesh deformation network to deform the mesh according to motion parameters, which is then mapped to the coordinate of Gaussians to adjust the associated Gaussian primitives. Moreover, we develop a learnable attribute deformation network with multiple \text{local attribute controllers} to refine appearance dynamics during inference. This design achieves photorealistic rendering quality (over 28 dB PSNR for novel poses and 32 dB PSNR for novel views) while remaining lightweight for mobile deployment.

\item \textbf{An End-to-end Monocular Telepresence System with SOTA Performance:} We develop a real-time system for monocular telepresence, as illustrated in Fig. \ref{fig:pipeline}. The sender-side module extracts human model parameters from a single monocular RGB camera for efficient transmission, while the receiver uses these parameters to drive the pre-built avatar, enabling high-fidelity rendering with a low end-to-end latency of $\thicksim$ 80 ms.

\item \textbf{Real-world Validation on Consumer Hardware:} We demonstrate the practicality of our system on a PC and a Meta Quest 3 \cite{quest3}. Our implementation supports real-time 3D teleconference on a single device and runs smoothly at a rate of $\thicksim$ 60 FPS, enabling real-time on-device immersive experience.
\end{itemize}

The remainder of this paper is organized as follows. Section \ref{sec:background} reviews the relevant background and related work. Section \ref{sec:system_design} presents the end-to-end design of our system and its main components. Section \ref{sec:reconstruction} elaborates on the implementation of our personalized 3DGS avatar reconstruction algorithm for amortization. Section \ref{sec:implementation} presents the implementation of software and hardware. In Section \ref{sec:evaluation}, we evaluate the system performance through extensive experiments. Finally, we conclude the paper and discuss future work in Section \ref{sec:conclusions}.

\section{Background and Related Work}\label{sec:background}

\textbf{3D Telepresence.} The goal of 3D telepresence is to create a compelling sense of ``co-presence'' by enabling remote individuals to experience interactions as if they were in the same physical space \cite{magicstream, project_straline, telealoha, monoport, monoportrtl, holoportation, forge4d, evagaussian}. A typical 3D telepresence system encompasses the real-time capture of immersive 3D content, its efficient transmission over networks, and photorealistic rendering on end-user devices. Early pioneering systems, such as Project Starline \cite{project_straline}, VirtualCube \cite{virtualcube}, and Holoportation \cite{holoportation}, demonstrate impressive capabilities using complex multi-camera setups with depth sensors to reconstruct users' 3D likenesses. 
However, the high costs associated with these systems (e.g., \$15,000 for Tele-Aloha \cite{telealoha} and \$24,999 for Project Starline) make them inaccessible to typical consumers.

To address the prohibitive bandwidth requirements of volumetric streaming, recent studies have explored semantic communication approaches. For instance, MagicStream \cite{magicstream} transmits compact semantic information, such as body keypoints and lighting characteristics, instead of raw 3D data, significantly reducing bandwidth consumption to under 1 Mbps \cite{magicstream}. Nevertheless, this method still relies on expensive multi-camera RGB-D capture setups (3 RGB-D cameras, each \$500) with precise calibration and synchronization. Moreover, the final visual quality depends heavily on the neural reconstruction model, which often fails to achieve full photorealism. In contrast, our work embraces the low-bandwidth principle of transmitting semantic features \cite{magicstream} but targets higher visual fidelity by driving a pre-built animatable 3DGS avatar. Notably, our system operates using a single monocular RGB camera (less than \$20), enabling real-time photorealistic telepresence on consumer-grade hardware.  Table \ref{tab:system_comparison} provides a detailed comparison of existing systems and our Mon3tr.

\textbf{3D Representation.} The evolution of 3D content representation has been driven by the pursuit of realism and efficiency. Traditional formats, such as point clouds, meshes, and voxels, provide explicit geometry but often require large data volumes to capture fine details, leading to high bandwidth consumption for streaming applications \cite{metastream, project_straline, farfetchfusion, virtualcube, holoportation}. Moreover, the inherent noise and artifacts in the reconstruction process can compromise the visual quality of these formats. The introduction of NeRF has marked a paradigm shift by enabling photorealistic novel view synthesis through an implicit continuous function modeled by an MLP \cite{nerf}. While NeRF achieves remarkable rendering quality, its extensive per-ray sampling results in computational costs that are prohibitive for real-time applications.
More recently, 3DGS has emerged as a superior alternative, combining the efficiency of explicit point-based rasterization with the visual fidelity of neural fields \cite{3dgs, 2dgs, dass, l3gs}. 3DGS represents a scene using a set of anisotropic 3D Gaussians, each characterized by properties such as position $\mathbf{\mu}$, covariance $\mathbf{\Sigma}$, color $\mathbf{c}$, and opacity $o$. These Gaussians are rendered efficiently via differentiable rasterization, where the final pixel color $C$ is computed through $\alpha$-blending over overlapping Gaussians $\mathcal{N}$ sorted by depth:
$$ C=\sum_{i\in \mathcal{N}}c_{i}\alpha_{i}\prod_{j=1}^{i-1}(1-\alpha_{j}). $$
This approach achieves real-time high-fidelity rendering but generates large scene files (often $>$ 1 GB), which poses significant challenges for network delivery \cite{l3gs}.

\begin{table*}[t!]
    \small
    \centering
    \caption{Comparison of Mon3tr with existing immersive telepresence systems \cite{holoportation, project_straline, farfetchfusion, monoport, monoportrtl, metastream, magicstream, telealoha}. }
    \label{tab:system_comparison}
    \setlength{\tabcolsep}{3pt}
    \begin{tabular*}{\linewidth}{@{\extracolsep{\fill}} c c c c c c c}
        \hline
        \textbf{System} & \textbf{\makecell{Full Body}} & \textbf{\makecell{Visual Quality}} & \textbf{Headset} & \textbf{\makecell{Bandwidth Requirement}} & \textbf{Monocular} & \textbf{Cost}\\
        \hline
        Holoportation \cite{holoportation} & \ding{51} & High & \ding{51} & High & \ding{55} & High\\
        Project Starline \cite{project_straline} & \ding{55} & Middle & \ding{55} & High & \ding{55} & High\\
        VirtualCube \cite{virtualcube} & \ding{55} & Middle & \ding{55} & High & \ding{55} & High\\
        FarfetchFusion \cite{farfetchfusion} & \ding{55} & Middle & \ding{55} & Middle & \ding{55} & Middle\\
        MonoPort \cite{monoport, monoportrtl} & \ding{51} & Middle & \ding{55} & Middle & \ding{51} & Low\\
        MetaStream \cite{metastream} & \ding{51} & Middle & \ding{51} & High & \ding{55} & High\\
        MagicStream \cite{magicstream} & \ding{51} & Middle & \ding{51} & Low & \ding{55} & High\\
        TeleAloha \cite{telealoha} & \ding{51} & High & \ding{55} & Middle & \ding{55} & High\\
        \textbf{Mon3tr (Ours)} & \textbf{\ding{51}} & \textbf{High} & \textbf{\ding{51}} & \textbf{Low} & \textbf{\ding{51}} & \textbf{Low}\\
        \hline
    \end{tabular*}
\end{table*}

\textbf{Photorealistic Avatar Reconstruction.} Creating photorealistic and animatable full-body avatars has long been a central challenge in computer graphics. A dominant paradigm is to deform a parametric template mesh, such as the Skinned Multi-Person Linear (SMPL) model \cite{smpl, smplx}, using linear blend skinning (LBS). While this effectively models basic body shapes and tight-fitting clothing, it struggles to capture complex geometry, such as hair, loose garments, and subtle surface details, thus limiting rendering quality \cite{arah, tava, avatarrex, monohuman}. To overcome these limitations, researchers have integrated 3DGS with parametric models. 
For example, 3DGS-Avatar \cite{3dgs_avatar} and Moreau et al. \cite{human_gs} use a single pose-conditioned MLP to model Gaussian deformations. However, these methods often fail to capture fine-grained details, such as facial expressions and clothing wrinkles, due to limited learning capacity. Subsequent approaches \cite{ash, uvgaussians, meshavatar} have leveraged convolutional neural networks (CNNs) to generate Gaussian property maps, thereby enhancing rendering quality. Further improvements like Animatable Gaussians \cite{animatable_gaussians} and DEGAS \cite{degas} use large CNN architectures for detailed appearance modeling, but suffer from slow inference speeds, impeding real-time use. 
Moreover, these methods struggle with non-rigid deformations, particularly for loose clothing and hair. 

Recent research has focused on learning pose-dependent correctives through lightweight networks. An effective strategy is to deform Gaussian properties using blendshape models \cite{taoavatar, slrf, gao, ma}, representing body or clothing shapes as linear combinations of 3DGS bases, akin to the SMPL model. While these methods achieve high-quality results at faster speeds, they are often limited to partial body avatars \cite{slrf, gao, ma} and can incur significant storage overhead, making them inefficient for mobile devices. Importantly, all these approaches reveal a fundamental trade-off: a single unified model often struggles to simultaneously capture intricate facial expressions and non-rigid body dynamics with high fidelity. To overcome this challenge, our Mon3tr adopts a ``divide and conquer'' strategy, integrating specialized modules for face, hands, body, and clothing. This modular design enables cohesive high-fidelity avatars without compromising details or real-time efficiency.

\section{System Design}
\label{sec:system_design}

The design of Mon3tr is guided by a philosophy of leveraging computational power to drastically reduce communication overhead, thereby enabling high-fidelity telepresence on consumer-grade hardware and networks. This section presents our system's core design philosophy and architecture, followed by detailed descriptions of the offline reconstruction process and the real-time online inference pipeline that enable our novel communication paradigm. The system architecture of Mon3tr and its workflow for capturing the 3DGS-based parametric human model are illustrated in Fig. \ref{fig:pipeline}.
\vspace{-3mm}

\subsection{Design Philosophy}
Immersive telepresence faces a fundamental trade-off between fidelity, latency, and bandwidth. 
Traditional volumetric streaming approaches \cite{metastream, holoportation, project_straline, monoport, monoportrtl, magicstream} require substantial bandwidth to maintain sufficiently high fidelity and low latency, failing to deliver satisfactory performance in bandwidth-constrained networks. Consequently, they have to compromise visual quality for real-time performance. In contrast, we invert this paradigm through three core principles:

\textbf{Computing for Communication.} Leveraging the increasing computational power of edge devices, we extract compact semantic representations of the user on the sender (e.g., a PC) to animate high-fidelity models on the receiver (e.g., mobile VR). This approach allows the sender to transmit motion instructions rather than frame-by-frame raw visual data, effectively eliminating the bandwidth bottleneck with local computational resources.

\textbf{Cost Amortization.} We decouple the computationally intensive task of avatar creation from real-time interaction. 
Specifically, we first build a personalized 3DGS avatar on the cloud server using a comprehensive offline reconstruction process to establish a detailed geometric and appearance prior, thereby bypassing expensive per-frame computations. 
With this pre-built avatar, subsequent live sessions only need to infer the human model parameters associated with the avatar to animate it. This strategy amortizes the upfront cost to ensure scalability and practicality over the lifetime of the avatar.

\textbf{Divide and Conquer.} To manage the complexity of full-body animation, we decompose the avatar into face, hands, and body. Then, specialized modules are employed to process pose detection and deformation for each part in parallel before integrating them into a unified parametric model, enhancing both tracking accuracy and system efficiency.

\subsection{System Architecture}
The Mon3tr architecture comprises three main components: a sender, e.g., a PC equipped with a monocular RGB camera, for capturing human subjects, a WebRTC-based data channel \cite{webrtc} for transmitting
extracted features and controlling bandwidth, and a mobile receiver for inference, rendering, and display of the pre-built 3DGS avatar. To be specific:

\begin{figure}[t]
    \centering       
    \includegraphics[width=\columnwidth]{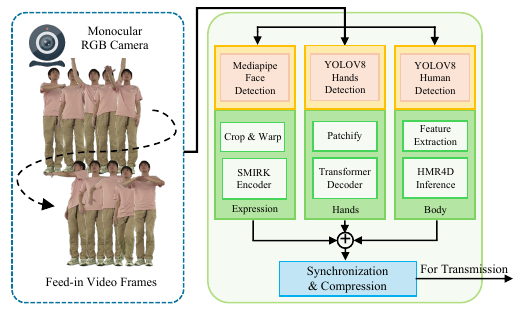} 
    \vspace{-20pt}
    \caption{Architecture of the sender pipeline.}
    \label{fig:sender}
\end{figure}

\begin{itemize}
\item \textbf{Sender (PC with a monocular RGB camera and a consumer-grade GPU):} 
The sender captures the poses and expressions of the user
using a RealSense RGB camera \cite{realsense} connected to a
PC equipped with an NVIDIA GeForce RTX 5090 D GPU.
To enhance the characterization of human motion, the PC runs a series of highly optimized pose
estimation models in parallel to extract compact parametric
representations of this user from the video feed, including
body poses $\boldsymbol{\theta}_{\text{b},t}$ \cite{smpl, gvhmr, nlf}, facial expressions $\boldsymbol{\psi}_t$ \cite{flame, smirk}, and hand gestures $\boldsymbol{\theta}_{\text{h},t}$ \cite{mano, hamer}, where $t$ is the frame index.
These parameters constitute the only data to be transmitted over networks. The architecture of the
sender is illustrated in Fig. \ref{fig:sender}.

\item \textbf{Communication channel:} We utilize WebRTC's \cite{webrtc} reliable data channel for low-latency transmission of the motion parameters. At the physical level, we use a GL.iNet GL-MT3000 router \cite{glinetmt3000} with ImmortalWrt firmware \cite{immortalwrt, openwrt} to manage the local
access network (LAN). We employ the Linux \textit{\textbf{tc}} command to simulate real-world bandwidth fluctuations and control user traffic.

\item \textbf{Receiver (VR headset):} The mobile receiver (e.g., Meta Quest 3 \cite{quest3}) downloads the pre-built user-specific avatar in the initialization phase. During online sessions, it receives the streams of unified motion parameters ($\boldsymbol{\theta}_{\text{b},t}$, $\boldsymbol{\psi}_t$, $\boldsymbol{\theta}_{\text{h},t}$) to drive the pre-built avatar. Specifically, we use two deformation networks $\mathcal{F}_{\text{mesh}}$ and $\mathcal{F}_{\text{attr}}$ to animate the avatar mesh and update Gaussian attributes, respectively, at the millisecond level for real-time rendering.

\end{itemize}

\begin{figure}[t]
    \centering
    \includegraphics[width=\columnwidth]{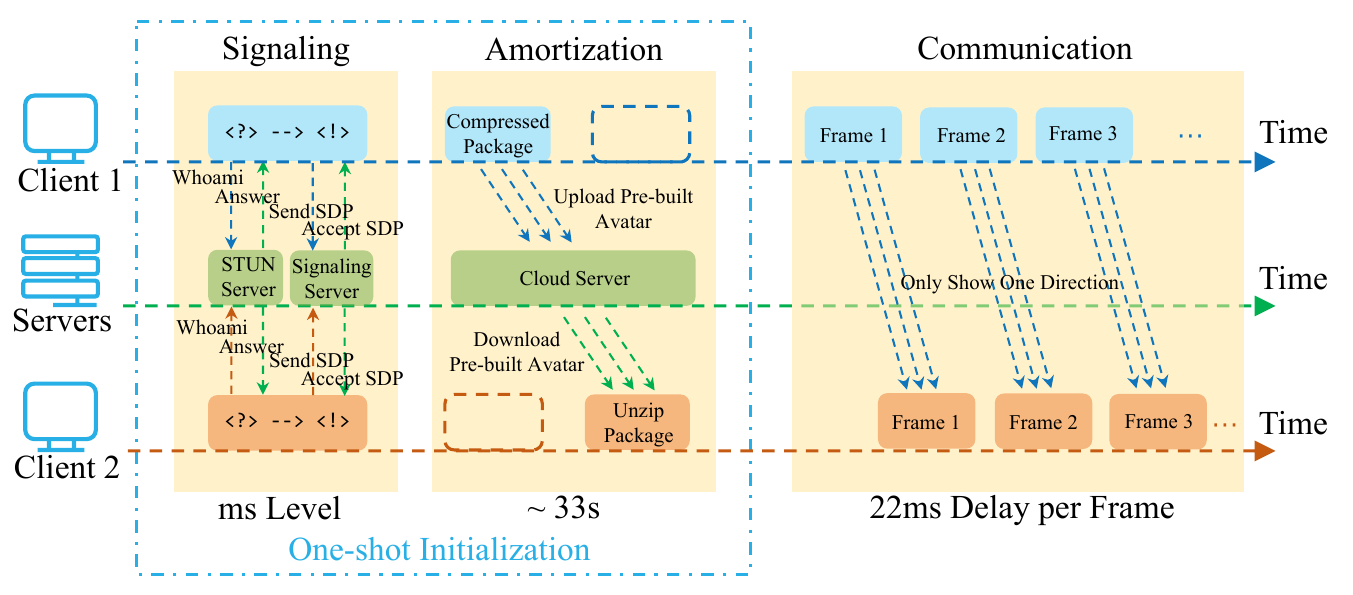}
    \caption{Network protocol.}
    \label{fig:protocol}
\end{figure}

This architecture minimizes network load by delegating photorealistic generation to the powerful computing capabilities of modern consumer endpoints. As shown in Fig. \ref{fig:protocol}, the setup of the network connection involves three stages: 1) \textit{Signaling}, which completes the handshake process at millisecond level; 2) \textit{Amortization}, which builds a user-specific avatar beforehand and saves it at the cloud server for subsequent downloading and usage, taking $\sim$ 33 s for one user; and 3) \textit{Communication}, which achieves data streaming with $\sim$ 22 ms delay for each frame, ensuring fluent immersive experiences.

\subsection{Offline Training} \label{sec:offline}
\begin{figure}[t]
    \centering     
    \includegraphics[width=\columnwidth]{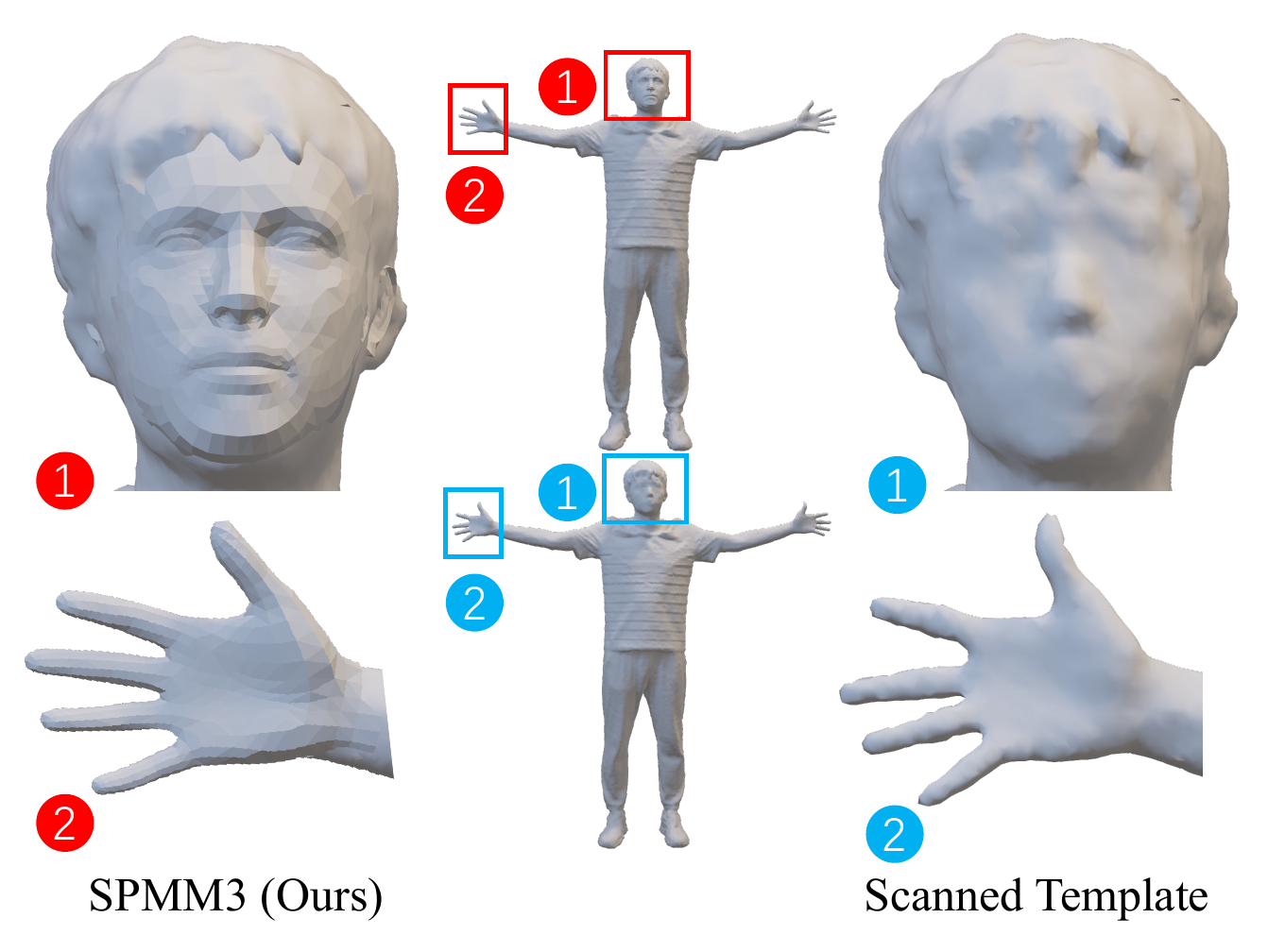}
    \caption{Comparison of different mesh templates.}
    \label{fig:template}
\end{figure}

The objective of this stage is to train a personalized, ultra-lightweight, and animatable avatar from a recorded multi-view RGB video of a user ($\sim$ 1-2 minutes in duration). This stage generates the core assets that are distributed to the receiver device prior to any communication session. The key steps are as follows:

\textbf{Parametric Body and Garment Modeling.} We introduce the \uline{s}kinned \uline{p}erson \uline{m}odel for \uline{M}on\uline{3}tr (\textbf{SPMM3}), a new parametric model to accurately capture personalized body shapes and loose clothing, which standard models like SMPL-X \cite{smplx} represent poorly. Specifically,
we construct a custom mesh from a tracked user scan \cite{physavatar} and rig it using skinning weight transfer \cite{skinweighttransfer}. However, this mesh struggles with non-rigid deformations. To address this limitation, we incorporate a mesh deformation network, denoted by $\mathcal{F}_{\text{mesh}}$, to learn these complex deformations. For the vertices $\mathbf{x}_\text{v}$ on the mesh, this network establishes a mapping $\mathcal{F}_{\text{mesh}}: (\boldsymbol{\theta}_{\text{b},t}, \boldsymbol{\psi}_t, \boldsymbol{\theta}_{\text{h},t}) \rightarrow \Delta \mathbf{x}_{\text{v}, t}$ at each frame $t$, where the pose and expression parameters ($\boldsymbol{\theta}_{\text{b},t}, \boldsymbol{\psi}_t, \boldsymbol{\theta}_{\text{h},t}$) are transformed into corrective vertex offsets $\Delta \mathbf{x}_{\text{v}, t}$, effectively capturing individual behavioral nuances. Moreover, for enhanced expressiveness, we seamlessly integrate pre-captured FLAME-based \cite{flame} face meshes and MANO-based \cite{mano} hand meshes into the base mesh at corresponding local positions. This integration significantly enriches the overall details of our model. In summary, SPMM3 offers a comprehensive solution for realistic body and garment representation as shown in Fig. \ref{fig:template}.

\textbf{Gaussian Binding and Corrective Model Learning.}
To achieve photorealism, we tightly bind 3D Gaussians to our personalized mesh and train an attribute deformation network $\mathcal{F}_{\text{attr}}$ that comprises multiple lightweight controllers to collaboratively learn corrections in avatar appearance. Specifically, each local attribute controller learns a set of corrective Gaussian properties, each of which we call a \textit{dragging force}. The total change of each Gaussian is then calculated as the weighted sum of all \textit{dragging forces}. This entire process is expressed as  $\mathcal{F}_{\text{attr}}: (\boldsymbol{\theta}_{\text{b},t}, \boldsymbol{\psi}_t, \boldsymbol{\theta}_{\text{h},t}) \rightarrow \Delta \mathbf{G}_{\text{attrs},t}$, 
where $\Delta \mathbf{G}_{\text{attrs},t} = \{\Delta \mathbf{x}_t, \Delta \mathbf{r}_t, \Delta \mathbf{s}_t, \Delta o_t, \Delta \mathbf{c}_t\}$ denotes the attribute deformations of position $\mathbf{x}$, rotation $\mathbf{r}$, scale $\mathbf{s}$, opacity $o$, and color $\mathbf{c}$ at frame $t$. 
We adapt the control strategy from \cite{mmlphuman, scgs} while incorporating an adaptive sampling scheme for control points. Guided by deformation analysis, this scheme enables us to sample points more densely in complex regions, such as the face, hands, and fabric wrinkles, enhancing the overall fidelity of our model's appearance corrections. 

\subsection{Online Inference}
The online inference stage constitutes the real-time telepresence session, which includes a sender-side process for parameter extraction and a receiver-side process for avatar reconstruction and rendering.

\textbf{Sender-side Process.}
To accurately capture fine details such as facial expressions and hand gestures from a single RGB camera, which is challenging for monocular capture, we use a concurrent estimation pipeline. First, a coarse full-body pass estimates the global pose $\boldsymbol{\theta}_{\text{b},t}$ \cite{gvhmr} and detects bounding boxes for the face and hands. These regions are then cropped and processed in parallel by specialized high-fidelity models such as HaMeR \cite{hamer} and SMIRK \cite{smirk}. A facial landmark detector yields expression parameters $\boldsymbol{\psi}_t$, while a hand model estimates pose parameters $\boldsymbol{\theta}_{\text{h},t}$. These parameters are fused with the global pose to produce the unified driving parameters of SPMM3 $\mathcal{P}_t \triangleq \{\boldsymbol{\theta}_{\text{b},t}, \boldsymbol{\psi}_t, \boldsymbol{\theta}_{\text{h},t}\}$, which are then compressed and transmitted to the receiver.

\textbf{Receiver-side Process.}
At the beginning of each session, the receiver fetches a personalized avatar package for each participant from a cloud repository. This package contains the template mesh $\mathbf{V}_\text{template}$, baseline Gaussian attributes $\mathbf{G}_\text{template}$, and pre-trained deformation models in ONNX format ($\mathcal F^{*}_\text{mesh}$, $\mathcal F^{*}_\text{attr}$) \cite{onnx}. Upon receiving the SPMM3 parameters at frame $t$, i.e., $\mathcal P_t = \{\boldsymbol{\theta}_{\text{b},t}, \boldsymbol{\psi}_t, \boldsymbol{\theta}_{\text{h},t}\}$, via the WebRTC data channel, a real-time on-device inference loop is executed. First, the lightweight mesh deformation network $\mathcal{F}^{*}_{\text{mesh}}$ applies corrective non-rigid offsets to $\mathbf{V}_\text{template}$. Next, the SPMM3 template mesh is posed using standard LBS with the incoming pose $\mathcal P_t$, yielding the final position of the mesh vertices at frame $t$ as $\mathbf{V}_{\text{final}, t} = \text{LBS} (\mathbf{V}_{\text{template}} + \mathcal{F}^{*}_{\text{mesh}}(\mathcal P_t))$. Concurrently, the attribute deformation network $\mathcal{F}^{*}_{\text{attr}}$ generates corrective Gaussian attributes $\Delta \mathbf{G}_{\text{attrs}, t}$ from $\mathcal P_t$ to update the baseline Gaussians $\mathbf{G}_\text{template}$. Finally, the resulting Gaussians $\mathbf{G}_{\text{attrs}, t}=\mathbf{G}_\text{template}+\Delta \mathbf{G}_{\text{attrs}, t}$ are rendered using our optimized 3DGS rasterization pipeline, enabling photorealistic real-time rendering on mobile NPUs and GPUs.

\section{Amortized 3DGS Avatar Reconstruction}
\label{sec:reconstruction}

As outlined in Section \ref{sec:offline}, the offline reconstruction stage is a one-time, computationally intensive process designed to create a high-fidelity, personalized, and animatable avatar. This section elaborates on the novel algorithmic pipeline developed for this purpose, which forms the foundation for the real-time performance of Mon3tr.

\subsection{High-fidelity Parametric Template Reconstruction}
\label{sec:mesh_reconstruction}

The realism of an animatable 3DGS avatar is fundamentally constrained by the quality and expressiveness of its underlying mesh template. Previous algorithms have often opted to bind Gaussians directly onto a standard SMPL-X mesh, such as GaussianAvatar and 3DGS-Avatar \cite{3dgs_avatar, gaussianavatar}. While straightforward, these methods inherit the limitations of the parametric model itself, much like a bald and naked human body, struggling to represent the complex geometry of loose garments or hair, and are susceptible to inter-penetration artifacts during complex motions, as discussed in \cite{physavatar}. Another line of methods, such as PhysAvatar \cite{physavatar}, track a mesh that is topologically identical to the subject to capture fine details. However, these techniques primarily focus on body and clothing reconstruction and may face challenges with full-body non-rigid deformations. Importantly, they also lack an inherent mechanism for precise control of facial expressions. To address these challenges and achieve superior fidelity, our proposed \textbf{SPMM3} model integrates high-precision part-specific templates with a learnable non-rigid deformation field. The construction process of SPMM3 is detailed below.

\textbf{Hybrid Topology Construction.} 
Instead of relying on a monolithic template, we construct the canonical topology of SPMM3, denoted by $\boldsymbol{M}_{\text{SPMM3}}$, by seamlessly fusing a personalized base body mesh with specialized high-fidelity components. We begin by obtaining a high-quality T-pose template mesh $\boldsymbol{M}_{\text{body}}$ of the subject via surface reconstruction \cite{pgsr}. To enhance expressiveness, we replace the coarse facial and hand regions of $\boldsymbol{M}_{\text{body}}$ with highly expressive FLAME-based face \cite{flame} and MANO-based hands \cite{mano}. 
Formally, the topology construction process is expressed as
\begin{equation}
    \boldsymbol{M}_{\text{SPMM3}} = \mathcal{U}\left( \boldsymbol{M}_{\text{body}}^{\text{masked}}, \mathcal{A}_\text{f}(\boldsymbol{M}_{\text{face}}(\boldsymbol{\beta}_\text{f}, \boldsymbol{\psi})), \mathcal{A}_\text{h}(\boldsymbol{M}_{\text{hand}}(\boldsymbol{\theta}_\text{h})) \right),
\end{equation}
where $\boldsymbol{M}_{\text{body}}^{\text{masked}}$ is the retopologized body mesh $\boldsymbol{M}_{\text{body}}$ with the regions of face and hands removed, $\boldsymbol{M}_{\text{face}}$ and $\boldsymbol{M}_{\text{hand}}$ are the FLAME-based face model and MANO-based hand model, respectively, with $\boldsymbol{\beta}_\text{f}$ representing the shape parameters of the face model, $\mathcal{U}$ represents replacing the face and hands of vanilla body mesh, $\mathcal{A}_\text{f}$ and $\mathcal{A}_\text{h}$ represent the rigid alignment transformations that register the face and hand components to the body kinematic chain. We then apply robust skinning weight transfer \cite{skinweighttransfer} to map the weights from the fitted SMPL-X model to this new hybrid topology $\boldsymbol{M}_{\text{SPMM3}}$, creating a unified, highly expressive armature ready for animation.

\textbf{Non-rigid Deformation Learning.} 
While $\boldsymbol{M}_{\text{SPMM3}}$ provides a superior static geometry, it lacks the ability to model non-rigid dynamics such as clothing wrinkles. To bridge this gap, we introduce a lightweight mesh deformation network $\mathcal{F}_{\text{mesh}}$ to predict vertex-wise offsets $\Delta \mathbf{x}_{\text{v}, t}$ in the canonical space. The final posed mesh $\boldsymbol{M}_{\text{posed}}$ is obtained via LBS:
\begin{equation}
    \boldsymbol{M}_{\text{posed}} = \text{LBS}\left( \boldsymbol{M}_{\text{SPMM3}} + \mathcal{F}_{\text{mesh}}(\mathcal P_t), \boldsymbol{\theta}_{\text{b},t}, \mathcal{J}(\boldsymbol{\beta}_\text{b}), \mathcal{W} \right),
\end{equation}
where $\mathcal{J}(\boldsymbol{\beta}_\text{b})$ denotes the joints' positions, $\boldsymbol{\beta}_\text{b}$ denotes the canonical body shape before skinning weight transfer, and $\mathcal{W}$ denotes the skinning weights of $\boldsymbol{M}_{\text{body}}$ after skinning weight transfer \cite{skinweighttransfer}.

\textbf{Training Objective.} 
The mesh deformation network $\mathcal{F}_{\text{mesh}}$ is supervised using a sequence of topologically consistent ground-truth meshes derived from a per-frame surface deformation pipeline \cite{physavatar}. We train the network by minimizing a composite loss function:
\begin{equation}
    \mathcal{L}_{\text{MLP}} = \mathcal{L}_{\text{vert}} + \lambda_{\text{n}} \mathcal{L}_{\text{normal}} + \lambda_{\text{nm}} \mathcal{L}_{\text{normal-map-f/b}},
\end{equation}
which comprises a vertex position $L_1$ loss $\mathcal{L}_{\text{vert}} = ||\mathbf{x}_{\text{MLP}} - \mathbf{x}_{\text{GT}}||_1$, a vertex normal consistency loss $\mathcal{L}_{\text{normal}} = ||\mathbf{n}_{\text{MLP}} - \mathbf{n}_{\text{GT}}||_1$, and a rendered normal map loss for the front and back views $\mathcal{L}_{\text{normal-map-f/b}}=\lambda_\text{r} \|\hat{\boldsymbol{I}}_{\text{n}} - \boldsymbol{I}_{\text{n}}\|_{1} + (1 - \lambda_\text{r}) \cdot \text{SSIM}(\hat{\boldsymbol{I}}_{\text{n}}, \boldsymbol{I}_{\text{n}})$. Here, $\mathbf{x}_{\text{MLP}}$ and $\mathbf{x}_{\text{GT}}$ are the predicted and ground-truth vertices, $\mathbf{n}_{\text{MLP}}$ and $\mathbf{n}_{\text{GT}}$ are the predicted and ground-truth normals, $\hat{\boldsymbol{I}}_{\text{n}}$ and $\boldsymbol{I}_{\text{n}}$ are the rendered and ground-truth normal maps, respectively.
This formulation yields a fully parametric photorealistic mesh $\boldsymbol{M}_{\text{posed}}$ that serves as a robust foundation for our subsequent Gaussian attribute learning.

\subsection{Distributed Local Gaussian Attribute Refinement}

While the SPMM3 mesh offers a geometric foundation, modeling the volumetric appearance requires handling complex dynamics. 
To improve the characterization of these dynamics, we train a 3DGS attribute deformation network $\mathcal{F}_{\text{attr}}$ to refine the Gaussian primitives binding to the mesh. This network comprises multiple \textit{local attribute controllers} that correct the attributes of Gaussians.
Instead of using standard attribute regression methods \cite{3dgs_avatar, meshavatar}, we conceptualize Gaussians as particles tethered within a \text{tension field} generated by a sparse set of controllers. These controllers exert ``\textit{dragging forces}'' on the Gaussians, effectively translating pose changes into coherent attribute deformations.

\textbf{Local Attribute Controllers and Potential Fields.} 
We instantiate a series of  \textit{local attribute controllers} $\{\Phi^j\}$ at uniformly sampled positions $\{\mathbf{y}_j\},j=1, \cdots, 500,$ on the canonical mesh surface $\boldsymbol{M}_{\text{body}}$. Each controller $\Phi^j$ functions as a local tension generator. It maps the input pose parameters $\mathcal{P}_t = \{\boldsymbol \theta_{\text{b},t}, \boldsymbol \psi_t, \boldsymbol \theta_{\text{h},t}\}$ to a high-dimensional vector $\mathbf{u}_{t}^j = \Phi^j(\mathcal{P}_t)$, which represents the tendency of appearance changes in the local manifold neighborhood at frame $t$ and is referred to as a \textit{displacement potential}.

\textbf{Virtual Mass and Dragging Force.} 
For a Gaussian $i$ tethered to the surface at position $\mathbf{x}_i$, the influence of a nearby controller $\Phi^j$ at position $\mathbf{y}_j$ is governed by a coupling strength, which we conceptualize as the \textit{virtual mass} $f$. To ensure spatial consistency and avoid pseudo-proximity artifacts (e.g., between the torso and arm), we define the virtual mass $f(\mathbf{x}_i, \mathbf{y}_j)$ by incorporating both geodesic distance $d_{\text{geo}}$ and skinning weight similarity $S_{\text{skin}}$ as
\begin{equation}
    f(\mathbf{x}_i, \mathbf{y}_j) = \frac{S_{\text{skin}}(\mathbf w_{\mathbf{x}_i}, \mathbf w_{\mathbf{y}_j)}}{d_{\text{geo}}(\mathbf{x}_i, \mathbf{y}_j) + \epsilon},
\end{equation}
where $\epsilon$ is a stability constant, $\mathbf w_{\mathbf{x}_i}$ and $\mathbf w_{\mathbf{y}_j}$ are the skinning weights of $\mathbf{x}_i$ and $\mathbf{y}_j$, respectively. 
Then, we calculate the \text{total dragging force} $F_{i,t}$ exerted on Gaussian $i$ by aggregating the displacement potentials from its $K=3$ nearest controllers, weighted by their respective virtual masses. This mirrors a force transfer operation:
\begin{equation} \label{force}
   F_{i,t} = \gamma_i\sum_{j \in \mathcal{N}_K(i)} f(\mathbf{x}_{i}, \mathbf{y}_j) \cdot \mathbf{u}_{t}^j,
\end{equation}
where $\gamma_i=\frac{1}{\sum_{j \in \mathcal{N}_K(i)} f(\mathbf{x}_{i}, \mathbf{y}_j)}$ represents the normalization term. The force $F_{i,t}$ does not directly displace the Gaussian but accumulates a potential that activates a pre-defined set of $B$ linear deformation bases $\{\Delta \mathbf{G}_{i}^b\}_{b = 1}^B$. The final attribute residuals $\Delta \mathbf{G}_{i,t}$ are obtained by projecting the force components onto these bases: $\Delta \mathbf{G}_{i,t} = \sum_{b=1}^{B} F_{i,t}[b] \cdot \Delta \mathbf{G}_{i}^b$. This mechanism allows the smoothly interpolated forces to drive highly detailed and non-linear attribute changes.

\begin{figure}[t]
    \centering 
    \includegraphics[width=\columnwidth]{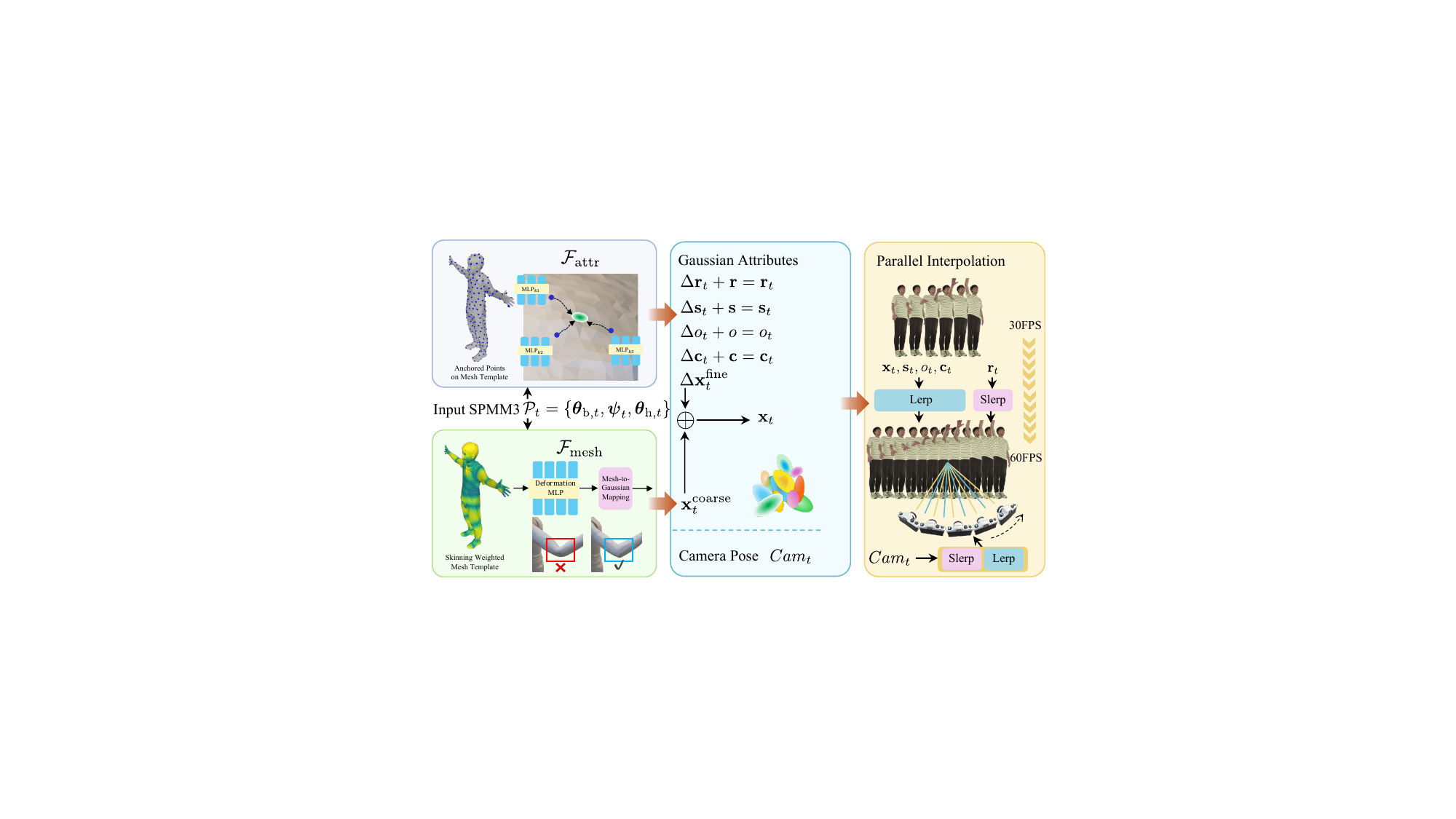}
    \caption{Inference data flow of 3DGS avatar animation.}
    \label{fig:inference}
\end{figure}

\textbf{Hierarchical Position Dynamics.} 
Crucially, to handle complex non-rigid motions, we model the Gaussian position $\mathbf{x}_{i,t}$ using a hierarchical formulation that decouples coarse deformations from fine-grained dynamics. The final position is modeled as:
\begin{equation} \label{eq:pos_update}
    \mathbf{x}_{i,t} = \mathbf{x}_{i,t}^{\text{coarse}} + \Delta \mathbf{x}_{i,t}^{\text{fine}}.
\end{equation}
Here, $\mathbf{x}_{i,t}^{\text{coarse}}=\mathcal{F}_{\text{m2g}}\left( \boldsymbol{M}_{\text{posed}} \right)$ represents the coarse motion of Gaussains, which is derived from the deformed mesh $\boldsymbol{M}_{\text{posed}}$ obtained in Section \ref{sec:mesh_reconstruction}, effectively transferring the non-rigid mesh deformations to the Gaussians using a mapping function $\mathcal{F}_{\text{m2g}}$. The second term, $\Delta \mathbf{x}_{i,t}^{\text{fine}}$, is predicted by the attribute deformation network $\mathcal{F}_{\text{attr}}$ to capture fine-grained surface residuals (e.g., cloth sliding) that the mesh resolution cannot resolve. The details of 3DGS avatar animation are demonstrated in Fig. \ref{fig:inference}.

\textbf{Optimization and Regularization.} 
We train the entire pipeline in an end-to-end manner with a total objective function $\mathcal{L}$ that balances photometric reconstruction with geometric regularization:
\begin{equation}
    \mathcal{L} = \mathcal{L}_{\text{photo}} + \lambda_{\text{g}}\mathcal{L}_{\text{guide}} + \lambda_{\text{s}}\mathcal{L}_{\text{scale}} + \lambda_{\text{b}}\mathcal{L}_{\text{bind}}.
\end{equation}
where the photometric loss $\mathcal{L}_{\text{photo}}$ itself is a combination of three metrics:
\begin{equation}
    \mathcal{L}_{\text{photo}} = (1-\alpha-\beta)\mathcal{L}_{1} + \alpha\mathcal{L}_{\text{SSIM}} + \beta\mathcal{L}_{\text{LPIPS}}.
\end{equation}
Here, $\mathcal{L}_{1} = \|\hat{\boldsymbol{I}} - \boldsymbol{I}\|_1$ is the $L1$ loss that encourages per-pixel color accuracy between the rendered and ground-truth images, $\mathcal{L}_{\text{SSIM}}$ is the structural similarity index measure (SSIM) that evaluates the similarity of structural information, and $\mathcal{L}_{\text{LPIPS}}$ is the learned perceptual image patch similarity (LPIPS) loss \cite{lpips} that captures human perceptual similarity. 
To prevent the local attribute controllers from predicting chaotic offsets, we enforce a \text{guide coherence loss} $\mathcal{L}_{\text{guide}}$ that penalizes divergence between neighboring controllers:
\begin{equation}
    \mathcal{L}_{\text{guide}} = \sum_{j,k = 1, \cdots, 500, j \neq k} \| \Delta \mathbf{x}_{\Phi^j} - \Delta \mathbf{x}_{\Phi^k}\|_2^2,
\end{equation}
where $\Delta \mathbf{x}_{\Phi^j}$ denotes the position offsets caused by the controller $\Phi^j$.
We also impose a scale regularization term
\begin{equation}
\mathcal{L}_{\text{scale}} = \sum_i \max(\|\mathbf{s}_i\| - s_{\text{thresh}}, 0)
\end{equation}
for Gaussians to prevent ``popping" artifacts, where $s_{\text{thresh}}$ is a pre-set threshold value. Finally, to prevent Gaussians from drifting away from the underlying geometry during complex articulation, we introduce a \textit{binding regularization} term:
\begin{equation}
    \mathcal{L}_{\text{bind}} = \sum_{i} \| \mathbf{x}_{i,t} - \text{Proj}_{\boldsymbol{M}_{\text{posed}}}(\mathbf{x}_{i,t}) \|_2^2,
\end{equation}
which penalizes the distance between each Gaussian and the nearest surface point on the deformed SPMM3 mesh $\boldsymbol M_{\text{posed}}$, ensuring that the volumetric avatar remains tightly coupled to the kinematic template.

\subsection{Inference on Mobile Devices}

\begin{figure}[t]
    \centering      
    \includegraphics[width=\columnwidth]{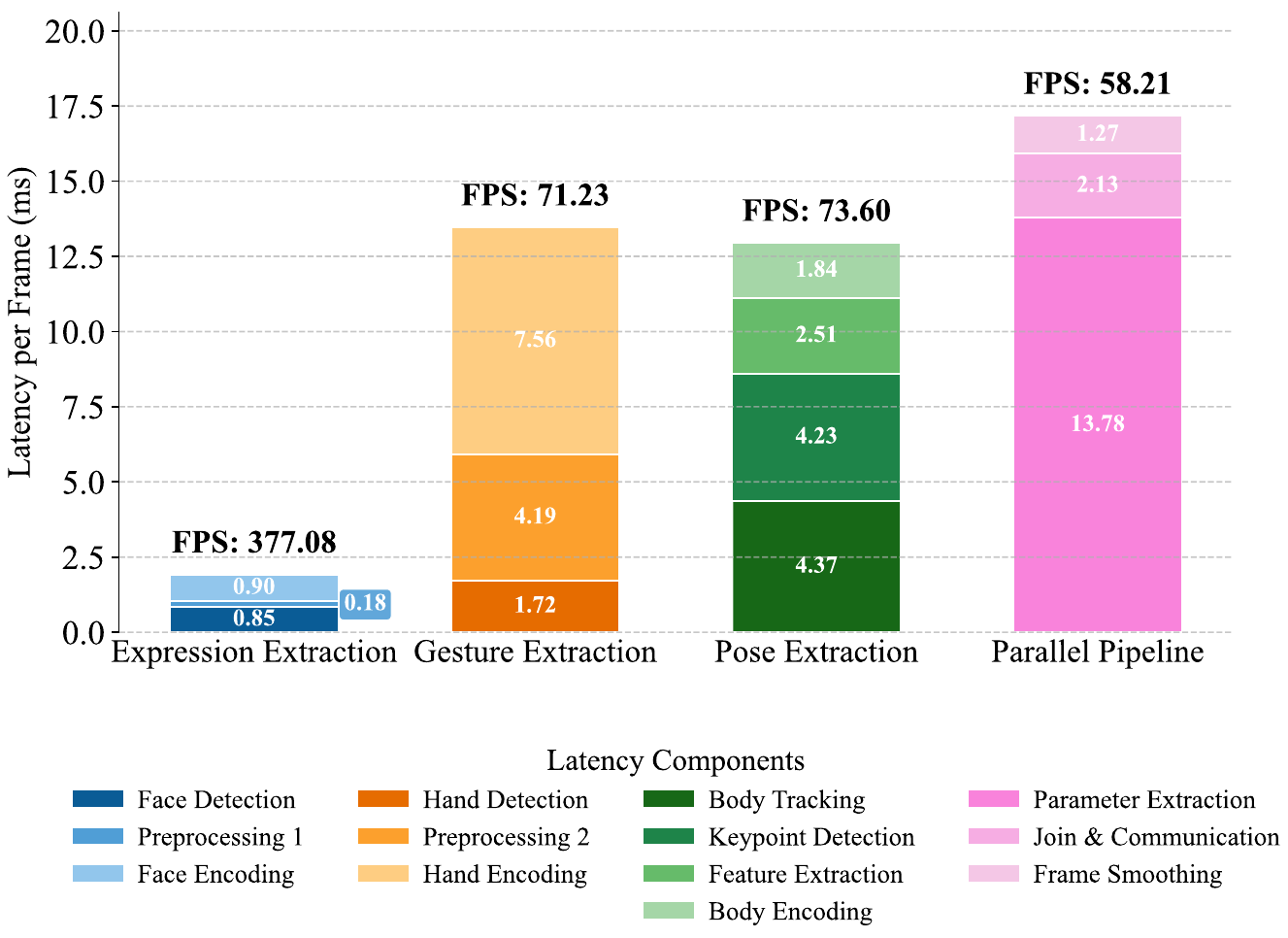}
    \caption{Latency breakdown of SPMM3 parameter extraction.}
    \label{fig:latency_breakdown}
\end{figure}

\textbf{Sender-side Capture.} On the sender side, we employ a suite of specialized models to capture the user's performance from RGB videos. Specifically, we extract body pose parameters $\boldsymbol \theta_{\text{b},t} \in \mathbb{R}^{75}$, facial expression coefficients $\boldsymbol\psi_t \in \mathbb{R}^{50}$, and hand gesture parameters $\boldsymbol \theta_{\text{h},t} \in \mathbb{R}^{90}$ at each frame $t$, respectively. The average per-frame latency for each component is detailed in Fig. \ref{fig:latency_breakdown}. These parameters are concatenated to form the set of SPMM3 motion parameters $\mathcal{P}_t$. To minimize the bandwidth requirement, this sequence is compressed using the FP16 quantization and lightweight LZ4 algorithm \cite{lz4}. The entire process of parameter extraction and transmission is optimized to run at a rate of $\sim$ 60 FPS, resulting in a low-bitrate stream of $<$ 0.2 Mbps, which makes it well-suited for real-time communication over conventional networks.

\textbf{Receiver-side Inference.} On the receiver side, two pre-trained lightweight deformation networks are deployed for real-time avatar animation: the mesh deformation network $\mathcal{F}_{\text{mesh}}$, which predicts non-rigid vertex offsets, and the 3DGS attribute deformation network $\mathcal{F}_{\text{attr}}$, which updates 3D Gaussian attributes. To ensure efficient inference on mobile hardware, both models are exported to the ONNX \cite{onnx} format as $\mathcal{F}^*_{\text{mesh}}$ and $\mathcal{F}^*_{\text{attr}}$, and further optimized for execution using the ONNX Runtime on Android devices. This optimization involves converting the models into fully static computation graphs, applying FP16 quantization to the deformation networks, utilizing UInt16 quantization for Gaussian sorting, and parallelizing frame interpolation to double the frame rate from 30 FPS to 60 FPS. This combination of techniques significantly accelerates inference while maintaining accuracy. The parallel interpolation process, depicted on the right hand side of Fig. \ref{fig:inference}, performs rotational interpolation using spherical linear interpolation (Slerp), while other attributes are interpolated linearly (Lerp). Consequently, the entire animation pipeline achieves a smooth streaming rate of $\sim$ 60 FPS on edge devices such as Meta Quest 3 \cite{quest3}. At the beginning of a communication session, these two user-specific models are downloaded from a cloud repository and loaded into the local environment to enable seamless avatar rendering.

\section{Implementation}\label{sec:implementation}

\textbf{Software.}
Our system utilizes a custom WebGL-based 3DGS renderer built on top of Three.js. At the sender, we employ a customized capture pipeline tailored for SPMM3. We adapt the capabilities of GVHMR \cite{gvhmr}, HaMeR \cite{hamer}, and SMIRK \cite{smirk} into a highly optimized workflow to extract holistic motion parameters in real time, including body pose $\boldsymbol \theta_{\text{b},t}$, hand gestures $\boldsymbol \theta_{\text{h},t}$, and facial expressions $\boldsymbol \psi_t$. To minimize data transmission, these parameters are compressed using the FP16 quantization and the lightweight LZ4 algorithm \cite{lz4}. All trained models, developed in PyTorch and PyTorch3D, are then exported to ONNX format \cite{onnx} to facilitate efficient inference and deployment on mobile devices.

\textbf{Hardware.}
The transmitter is a PC equipped with an NVIDIA GeForce RTX 5090 D GPU, which captures RGB video streams via an Intel RealSense camera \cite{realsense}. The receiver is a standalone Meta Quest 3 headset \cite{quest3} powered by a Snapdragon XR2 Gen 2 processor. To simulate realistic network conditions, we deploy a GL.iNet GL-MT3000 router \cite{glinetmt3000} with ImmortalWrt \cite{immortalwrt} to create isolated LANs, where the bandwidth is controlled using the Linux \textit{\textbf{tc}} command \cite{tc} to emulate real-world network fluctuations.

\section{Evaluation}\label{sec:evaluation}

\subsection{Experimental Setup}

\begin{figure}[t]
    \centering     
    \includegraphics[width=\columnwidth]{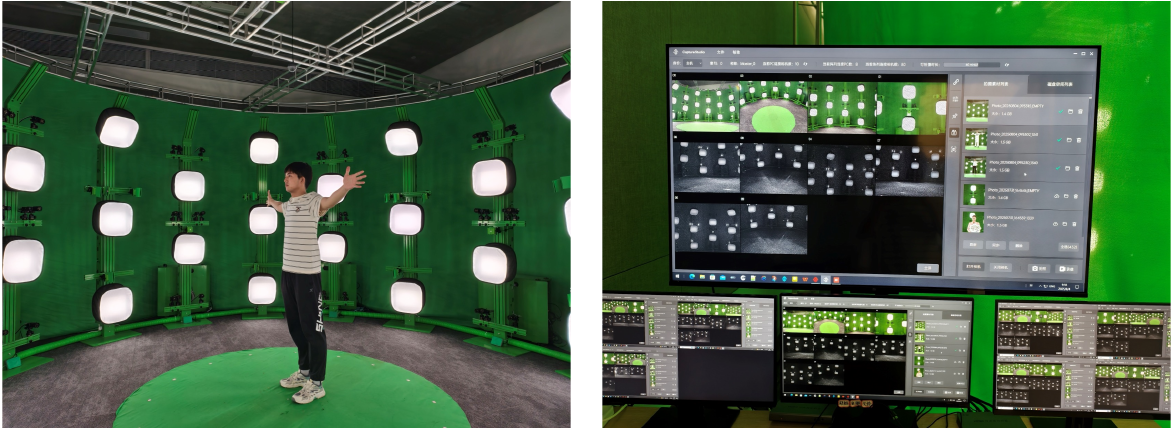} 
    \caption{iCom4D dataset collected by 32 synchronized 12MP cameras.}
    \label{fig:capture}
    \vspace{-10pt}
\end{figure}

\textbf{Data Collection.} To address the scarcity of comprehensive datasets for evaluating full-body talking avatars with nuanced facial expressions like ActorsHQ and AvatarRex, we introduce the iCom4D dataset. This dataset is specifically designed to benchmark performance in telepresence scenarios that involve rich body language and detailed facial articulation. It comprises multi-view recordings of two subjects captured by 32 synchronized 12MP cameras \cite{dreams} as shown in Fig. \ref{fig:capture}. Each subject performs two distinct sequences featuring a wide variety of large body poses and expressive facial movements, ensuring thorough coverage of motions. 

\begin{figure*}[t!]
    \centering
    \captionsetup{justification=centering}
    \includegraphics[width=1\textwidth]{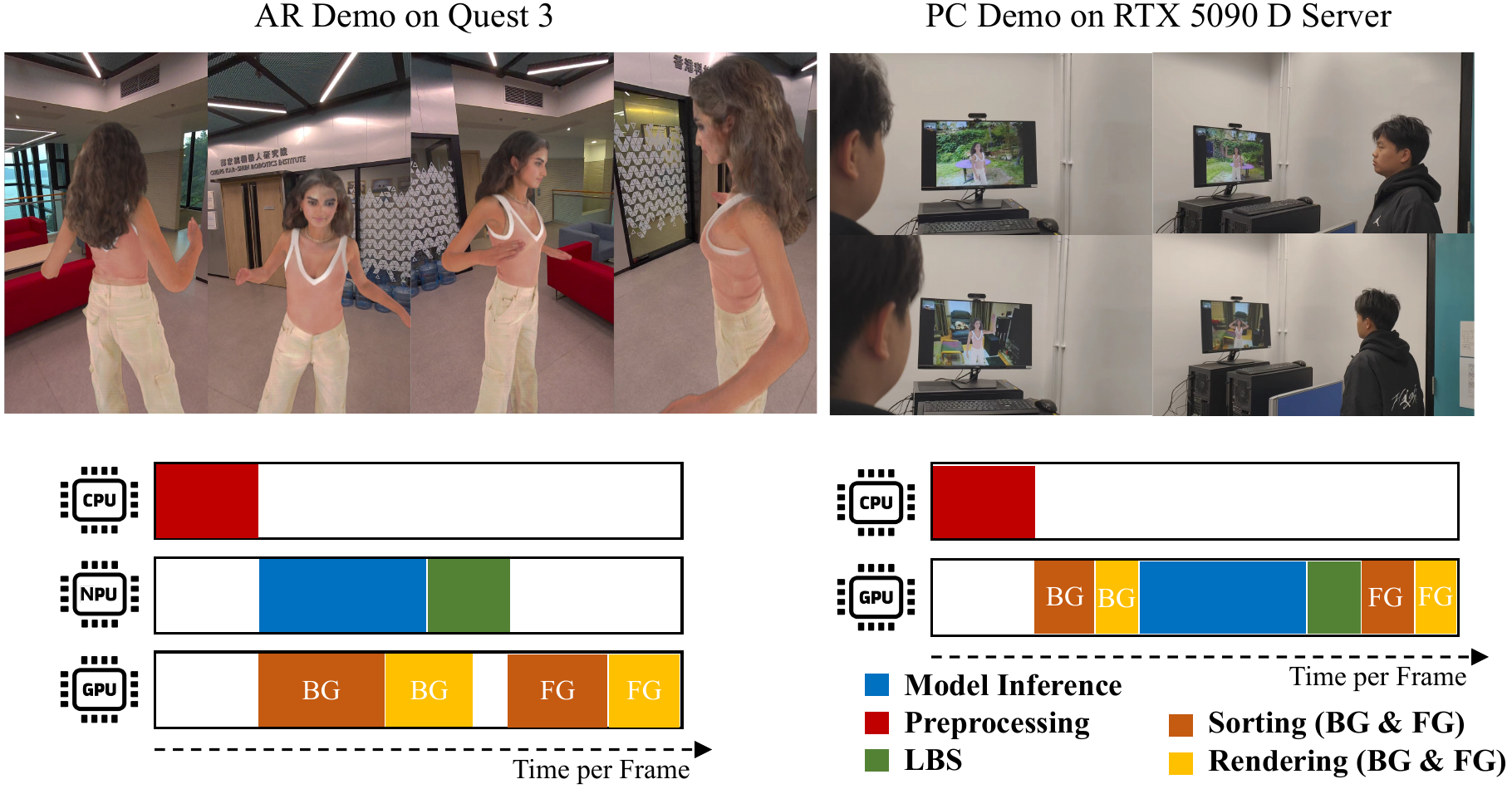}
    \caption{Resource allocation and live demo on Meta Quest 3 and consumer-level PC. BG and FG represent the background scene and foreground human, respectively.}
    \label{fig:results2}
\end{figure*}

\begin{figure}[t!]
    \centering
    \includegraphics[width=1\columnwidth]{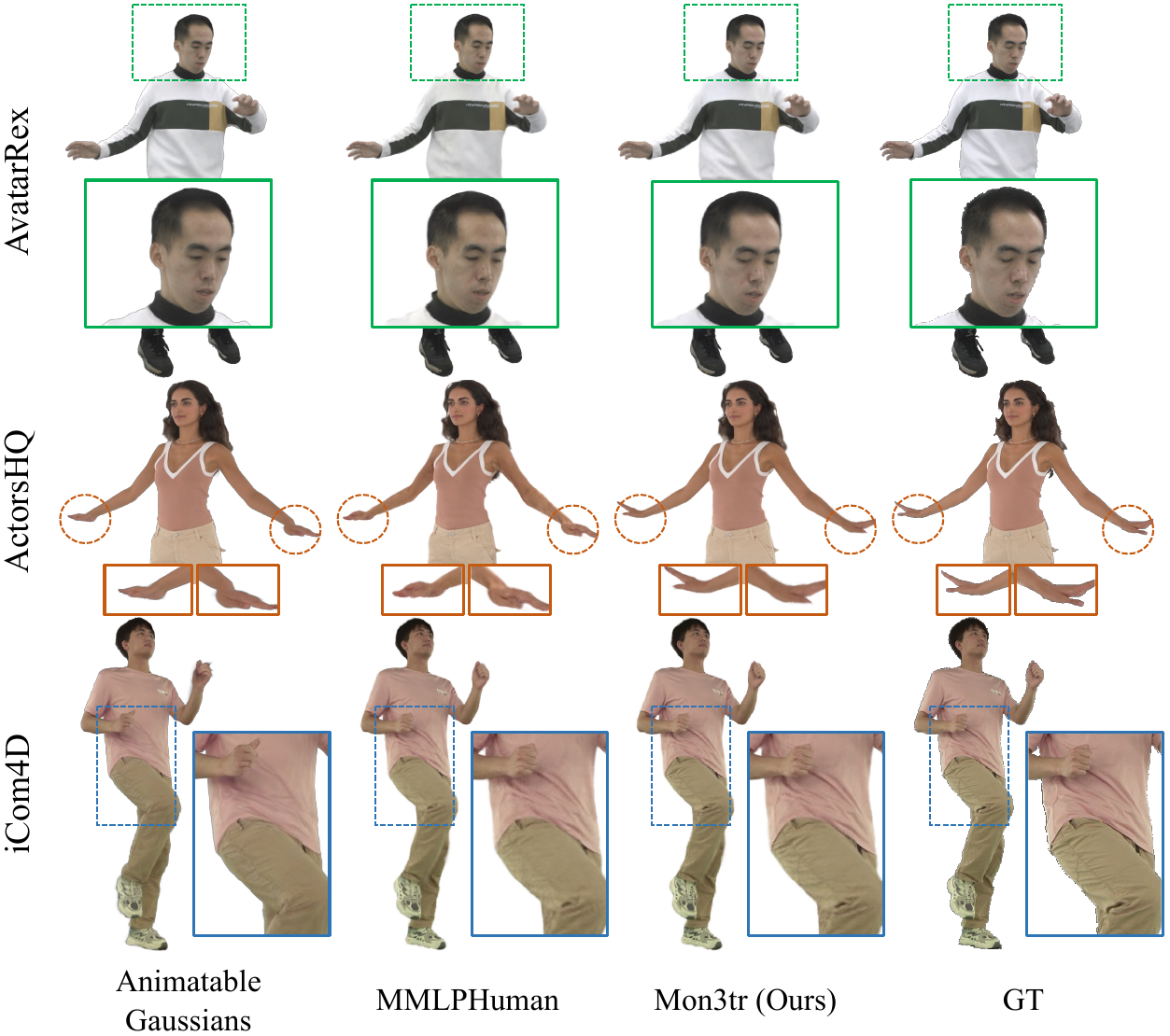}
    \caption{Qualitative comparison on the AvatarRex, ActorsHQ, iCom4D datasets.}
    \label{fig:results3}
\end{figure}

\textbf{Baselines.}
To ensure a comprehensive evaluation, we compare Mon3tr against state-of-the-art (SOTA) methods in two key areas: end-to-end system performance and separate evaluation of different components.
For system-level metrics, we consider closed-source methods, including MagicStream \cite{magicstream}, MetaStream \cite{metastream}, and TeleAloha \cite{telealoha}, and the open-source method MonoPort \cite{monoport, monoportrtl}. MagicStream serves as a benchmark for semantic communication efficiency, while MetaStream, TeleAloha, and MonoPort represent recent volumetric streaming systems. Since MonoPort only implements the system on a local server, we re-implement the system on our RTX 5090 D server to take into account the communication overhead and to make sure that the baseline is measured in the same hardware environment. This comparison focuses on end-to-end latency, bandwidth overhead, and overall FPS.
For SPMM3 motion parameter capture, we breakdown the latency of the three parallel algorithms for body motion, hand gesture, and facial expression extraction, respectively, and measure the overall latency when running them together. 
For visual rendering quality, we compare against leading animatable avatar creation methods: Animatable Gaussians \cite{animatable_gaussians}, MMLPHuman \cite{mmlphuman, scgs} and MeshAvatar \cite{meshavatar}. These methods represent the SOTA in high-quality human avatar modeling. We evaluate the visual quality of novel poses and views using standard metrics: PSNR, SSIM, LPIPS, and Fréchet inception distance (FID).

\textbf{Network Setup.} To control bandwidth between two clients on the same local network and to simulate real-world network conditions, we implement logical network separation on a single ImmortalWrt-based \cite{openwrt, immortalwrt} GL.iNet GL-MT3000 router \cite{glinetmt3000}. By default, intra-LAN traffic is managed by Layer 2 switching, bypassing the Layer 3 routing engine where queuing disciplines operate. Our setup forces traffic between the two target clients through the router's CPU, enabling network control via the Linux \textit{\textbf{tc}} command \cite{tc}. Specifically, we configure two separate network interfaces with non-overlapping subnets. Client A (wired, a PC with a monocular RGB camera) is connected to a dedicated LAN port assigned exclusively to a new bridge interface, \textit{Net\_A}, configured with the subnet $192.168.8.0/24$. Client B (wireless, a VR headset or mobile phone) connects to a dedicated SSID that is exclusively bridged to a second new interface, \textit{Net\_B}, with the subnet $192.168.9.0/24$. Firewall rules are established to allow forwarding between these two zones. Consequently, the traffic from Client A to Client B must egress through the \textit{Net\_B} interface, while the traffic from Client B to Client A must egress through the \textit{Net\_A} interface. This setup enables precise bidirectional bandwidth control by applying separate \textit{\textbf{tc}} egress rules to each interface. By replaying pre-recorded bandwidth traces on these two interfaces, we evaluate the communication effectiveness of our system under various real-world network conditions.

\textbf{Resource Allocation.}
We further evaluate the system's adaptability by analyzing resource scheduling across different hardware architectures, as visualized in Fig. \ref{fig:results2}. This comparison illustrates our tailored optimization strategies for different computing environments. In the mobile scenario, deployed on a Meta Quest 3, a significant contribution of our work is the development of a custom mobile-native 3DGS renderer designed specifically for heterogeneous computing. By implementing a strategic scheduling pipeline, we effectively offload computationally intensive tasks, specifically network inference and LBS, to the device's neural processing unit (NPU). This allows the GPU to focus exclusively on high-bandwidth tasks, such as sorting and rendering background (BG) and foreground (FG) elements, while the CPU manages preprocessing. This efficient task distribution is critical for ensuring real-time performance on resource-constrained mobile hardware. In contrast, in the PC setup equipped with an NVIDIA RTX 5090 D, where computational resources are abundant, the GPU manages the entire pipeline. Leveraging this performance headroom, we introduce a specialized eye-tracking module that dynamically adjusts the camera perspective based on the user's gaze, facilitating seamless interaction from varying viewing angles. This dual approach highlights our system's versatility: aggressively optimizing low-level efficiency via NPU offloading for mobile VR, while exploiting desktop capabilities to enhance interactive immersion.

\subsection{Component-Wise Evaluation}
\textbf{SPMM3 Parameter Extraction.} 
The real-time performance of our Mon3tr framework critically depends on the efficiency of the sender-side pipeline, which extracts comprehensive driving parameters from a single monocular RGB video stream. To evaluate this efficiency, we conduct a detailed latency breakdown of our parallel processing architecture, with results presented in Fig.~\ref{fig:latency_breakdown}. Our pipeline concurrently executes three specialized models for facial expressions, hand gestures, and body pose.
The results show that each component demonstrates high efficiency. Facial expression extraction achieves an impressive 377.08 FPS, with minimal latency. Body pose and hand gesture extractions operate at 73.60 FPS and 71.23 FPS, respectively. While hand gesture extraction incurs the highest latency, primarily due to its complex hand detection and inference stages, its performance remains well within real-time constraints.
When integrated into our parallel pipeline, the system achieves a final synchronized output rate of 58.21 FPS. The dominant latency component is ``Worker Execution'' (13.78 ms), which represents the concurrent operation of the three models. The overhead for output synchronization (``Join \& Communication'') and subsequent temporal smoothing is minimal, at 2.13 ms and 1.27 ms, respectively. This end-to-end throughput of nearly 60 FPS confirms that our sender-side process is capable of capturing and transmitting fluid, expressive human motion with sufficiently low latency to enable a seamless and immersive real-time telepresence experience.

\begin{table}[t!]
\centering
\footnotesize
\caption{Quantitative comparison with SOTA methods on training poses.}
\label{tab:render_table1}
\begin{tabular}{l@{\hspace{0.04em}} S[table-format=2.4]
                 S[table-format=1.4]
                 S[table-format=1.4]
                 S[table-format=2.4]}
\toprule
\textbf{Method} & \textbf{PSNR (dB)$\uparrow$} & \textbf{SSIM$\uparrow$} & \textbf{LPIPS$\downarrow$} & \textbf{FID$\downarrow$} \\
\midrule
MeshAvatar          & 28.9698          & 0.9527          & 0.0397          & 24.3030 \\
Animatable Gaussians & 31.1634          & 0.9770          & 0.0311          & 14.3777 \\
MMLPHuman    & 32.2833 & 0.9843 & 0.0240 & 13.1455 \\
\textbf{Mon3tr (Ours)}       & \textbf{32.4037} & \textbf{0.9857} & \textbf{0.0232} & \textbf{11.2907} \\
\bottomrule
\end{tabular}
\end{table}

\begin{table}[t!]
\centering
\footnotesize
\caption{Quantitative comparison with SOTA methods on novel poses.}
\label{tab:render_table2}
\begin{tabular}{l@{\hspace{0.3em}} S[table-format=2.4]
                 S[table-format=1.4]
                 S[table-format=1.4]
                 S[table-format=2.4]}
\toprule
\textbf{Method} & {\textbf{PSNR (dB)$\uparrow$}} & {\textbf{SSIM$\uparrow$}} & {\textbf{LPIPS$\downarrow$}} & {\textbf{FID$\downarrow$}} \\
\midrule
MeshAvatar & 27.2908 & 0.9626 & 0.0748 & 28.1674 \\
Animatable Gaussians & 27.5837 & 0.9710 & 0.0652 & 23.9481 \\
MMLPHuman & 28.3435& 0.9731 & 0.0606 & 21.1214 \\
\bfseries \textbf{Mon3tr (Ours)} & \textbf{28.3858} & \textbf{0.9743} & \textbf{0.0564} & \textbf{20.3608} \\
\bottomrule
\end{tabular}
\end{table}

\begin{table}[t!]
\centering
\footnotesize
\caption{Ablation study.}
\label{tab:render_table3}
\setlength{\tabcolsep}{4pt} 
\begin{tabular}{l@{\hspace{0.04em}} S[table-format=2.4]
                 S[table-format=1.4]
                 S[table-format=1.4]
                 S[table-format=2.4]}
\toprule
\textbf{Method} & {\textbf{PSNR (dB)$\uparrow$}} & {\textbf{SSIM$\uparrow$}} & {\textbf{LPIPS$\downarrow$}} & {\textbf{FID$\downarrow$}} \\
\midrule
w/o Mixted Template & 32.1492 & 0.9836 & 0.0265 & 12.6781 \\
w/o Weighted Interpolation & 32.0432 & 0.9828 & 0.0328 & 14.7563 \\
\bfseries \textbf{Full Design} & \textbf{32.4037} & \textbf{0.9857} & \textbf{0.0232} & \textbf{11.2907} \\
\bottomrule
\end{tabular}
\end{table}

\begin{figure*}[t!]
    \centering
    \captionsetup{justification=centering}
    \includegraphics[width=0.9\textwidth]{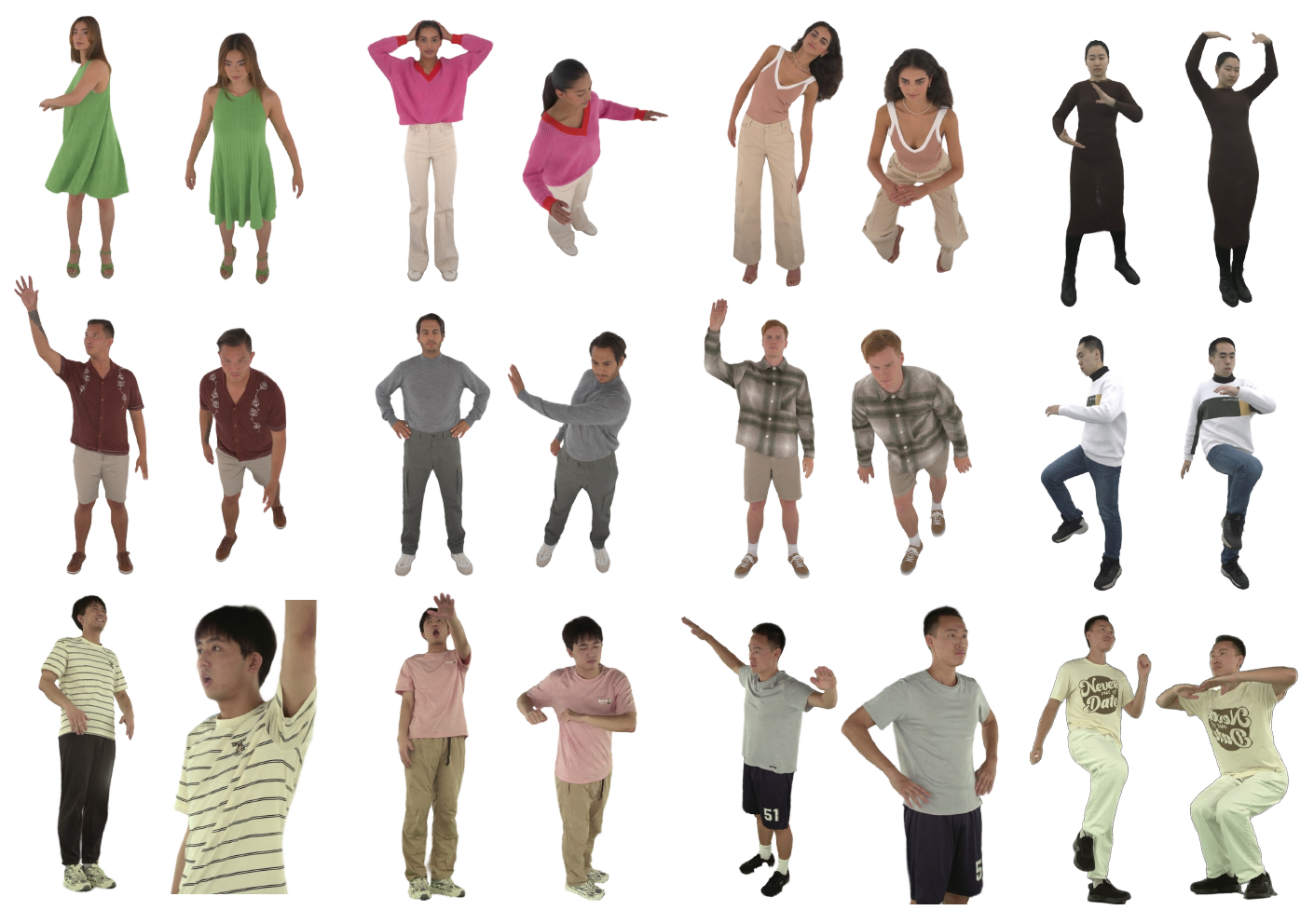}
    \caption{Qualitative results on ActorsHQ, AvatarRex, and iCom4D datasets.}
    \label{fig:results1}
\end{figure*}

\textbf{3DGS Avatar Rendering Quality.} 
To rigorously evaluate the visual fidelity of our generated avatars, we conduct quantitative comparisons against several SOTA methods, including MeshAvatar \cite{meshavatar}, Animatable Gaussians \cite{animatable_gaussians}, and MMLPHuman \cite{mmlphuman}. We assess performance under two distinct conditions: rendering quality on poses seen during training and, more critically, generalization capability on novel, unseen poses. We follow the same evaluation settings as those used in Animatable Gaussians \cite{animatable_gaussians}.
As detailed in Table \ref{tab:render_table1}, our method demonstrates superior performance on training poses. It achieves the highest scores across all metrics, with a PSNR of 32.4037 dB and an SSIM of 0.9857, while also attaining the lowest LPIPS of 0.0232 and FID of 11.2907. These results indicate that our model reconstructs the subject appearance with exceptional accuracy and photorealism for familiar motions.
Furthermore, the results for novel poses, presented in Table \ref{tab:render_table2}, highlight the robust generalization of our model. In this more challenging scenario, our method also outperforms all baseline models, achieving a leading PSNR of 28.3858 dB and the best scores for SSIM, LPIPS, and FID. This strong performance validates the effectiveness of our hybrid model, confirming its suitability for generating high-quality dynamic avatars in real-time telepresence applications. Furthermore, we validate the effectiveness of the two main components in the avatar learning framework in Table \ref{tab:render_table3}, where both components contributes to high-fidelity results. In addition, more qualitative results tested on the ActorsHQ, AvatarRex, and iCom4D datasets are demonstrated in Figs. \ref{fig:results3} and \ref{fig:results1}, which further shows high-fidelity rendering capability across different identities.

\subsection{End-to-End Evaluation}

\begin{table*}[t]
    \small
    \centering
    \caption{Quantitative comparison with existing telepresence systems.}
    \label{tab:table5}
    \setlength{\tabcolsep}{3pt}
    \begin{tabular*}{\linewidth}{@{\extracolsep{\fill}} l c c c c c}
        \hline
        \textbf{\makecell{System}}  & \textbf{\makecell{Test Device}} & \textbf{Camera Setup} & \textbf{Estimated Price (\$)} & \textbf{\makecell{Test Computational Platform}} & \textbf{VRAM Usage}\\
        \hline
        
        \multicolumn{2}{l}{\textit{Closed-source}} \\
        \quad MetaStream \cite{metastream} 
         & HoloLens 2 &  4 × RGBD & 167.5 & 4 × Jetson Nano + RTX 2080S & \textbackslash\\
        \quad MagicStream \cite{magicstream} 
         & HoloLens 2 &  3 × RGBD & 94.4 & RTX 4090 & 7.6 GB\\
        \quad TeleAloha \cite{telealoha} 
         & Stereo Screens &  4 × RGB & 149.0 & RTX 4090 & \textbackslash\\
        \hline
        
        \multicolumn{2}{l}{\textit{Open-source \& Ours}} \\
        \quad MonoPort \cite{monoport, monoportrtl} 
         & Dell Screens  & 1 × RGB & 89.1 & RTX 5090 D & 11.2 GB\\        
        \quad \textbf{Mon3tr (Ours)}  & \textbf{Quest 3 or Dell Screens} & \textbf{1 × RGB} & \textbf{73.1} & \textbf{RTX 5090 D} & \textbf{3.9 GB}\\
        \hline
        
    \end{tabular*}
\end{table*}

In our end-to-end evaluation, we benchmark Mon3tr against a comprehensive suite of leading telepresence systems, ranging from high-end closed-source solutions to accessible open-source frameworks. Our assessment focuses on four critical performance pillars that define the viability of immersive communication: hardware accessibility, bandwidth efficiency, interactive latency, and runtime rendering fluidity. Instead of relying on laboratory-grade workstation clusters or complex multi-sensor arrays, which are common in baseline methods, our experiments are strictly conducted on consumer-level hardware, pairing a standard PC sender with a standalone Meta Quest 3 receiver. This setup is designed to rigorously test the system's performance in real-world deployment scenarios. The following quantitative analyses demonstrate how our proposed amortized computation paradigm successfully disrupts the traditional trade-offs in telepresence, enabling high-fidelity 3D interaction without the prohibitive computational costs and network overheads that characterize conventional volumetric streaming approaches.

\textbf{Hardware and Cost Efficiency.} 
As detailed in Table \ref{tab:table5}, Mon3tr fundamentally lowers the barrier to entry for high-fidelity immersive telepresence. Unlike closed-source baselines such as TeleAloha or Project Starline, which necessitate expensive multi-camera arrays and high-end server clusters, our framework is validated on accessible consumer-grade hardware. Using a single monocular RGB camera and a standard GPU, we reduce the estimated system cost to approximately \$73. Furthermore, Mon3tr demonstrates exceptional memory efficiency, requiring only 3.9 GB of VRAM, which is significantly less than the 11.2 GB used by MonoPort. This confirms that our amortization strategy successfully decouples visual quality from hardware complexity, making photorealistic 3D communication feasible on ubiquitous personal devices.

\begin{figure}[t]
    \centering      
    \includegraphics[width=1\columnwidth]{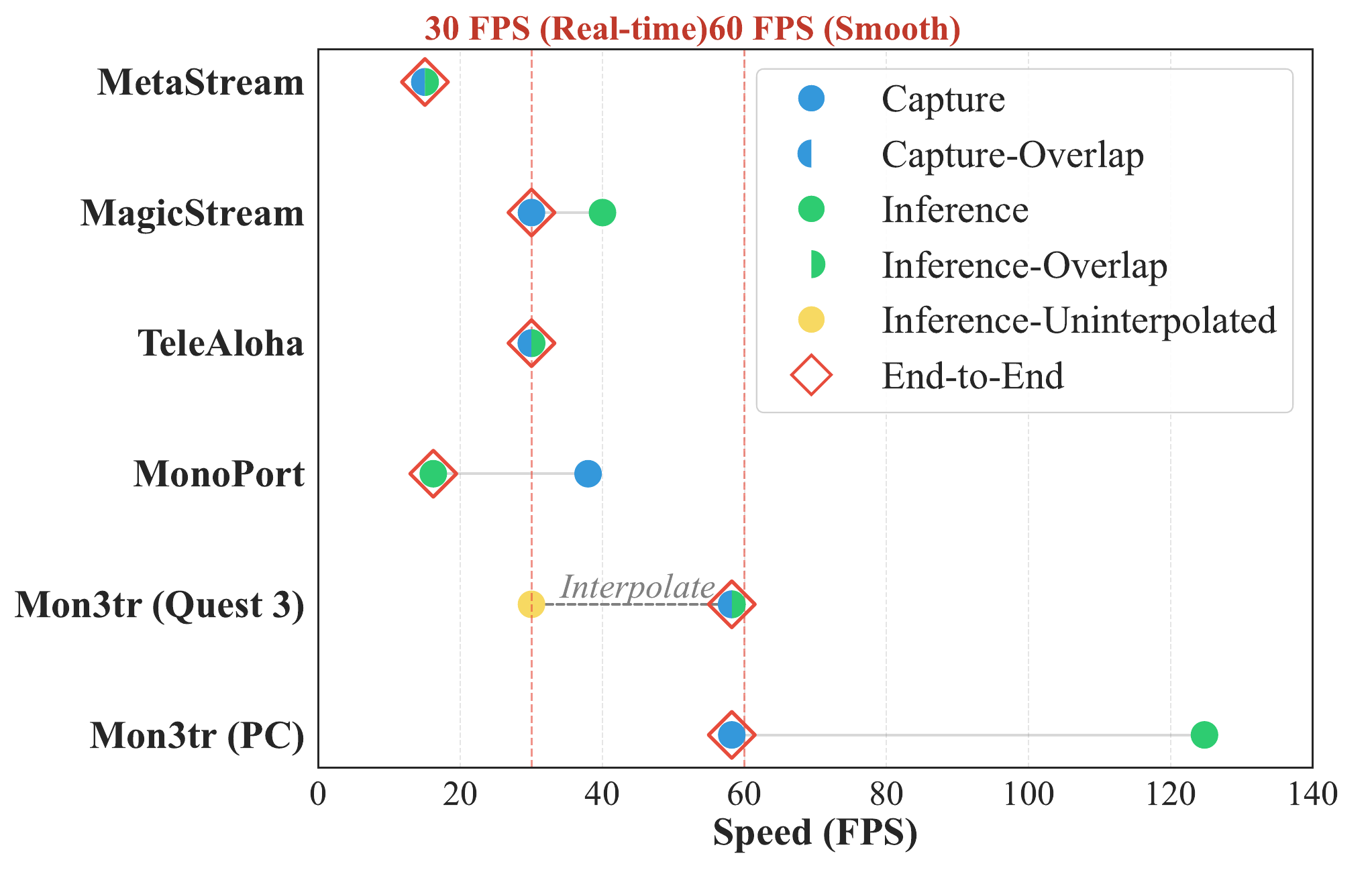}
    \caption{Comparison of system speed between Mon3tr and baselines, where the unit for the values is FPS.}
    \label{fig:fps}
\end{figure}

\textbf{Runtime FPS Analysis.} 
Fluid motion is paramount for preserving non-verbal cues in telepresence. Fig. \ref{fig:fps} benchmarks the runtime throughput, where Mon3tr establishes a new performance standard. Our system sustains a synchronized end-to-end frame rate of $\sim$ 60 FPS on the standalone Meta Quest 3, effectively doubling the 30 FPS performance of MagicStream and TeleAloha, and quadrupling the 15 FPS of MetaStream. On PC platforms, our lightweight inference engine achieves over 124 FPS, highlighting the efficiency of our parallelized SPMM3 parameter extraction and NPU-offloaded rendering. This performance gap validates that our lightweight deformation network based decoding can support cinematic realism in real time without the computational bottlenecks typical of traditional volumetric rendering.

\begin{figure}[t]
    \centering     
    \includegraphics[width=0.95\columnwidth]{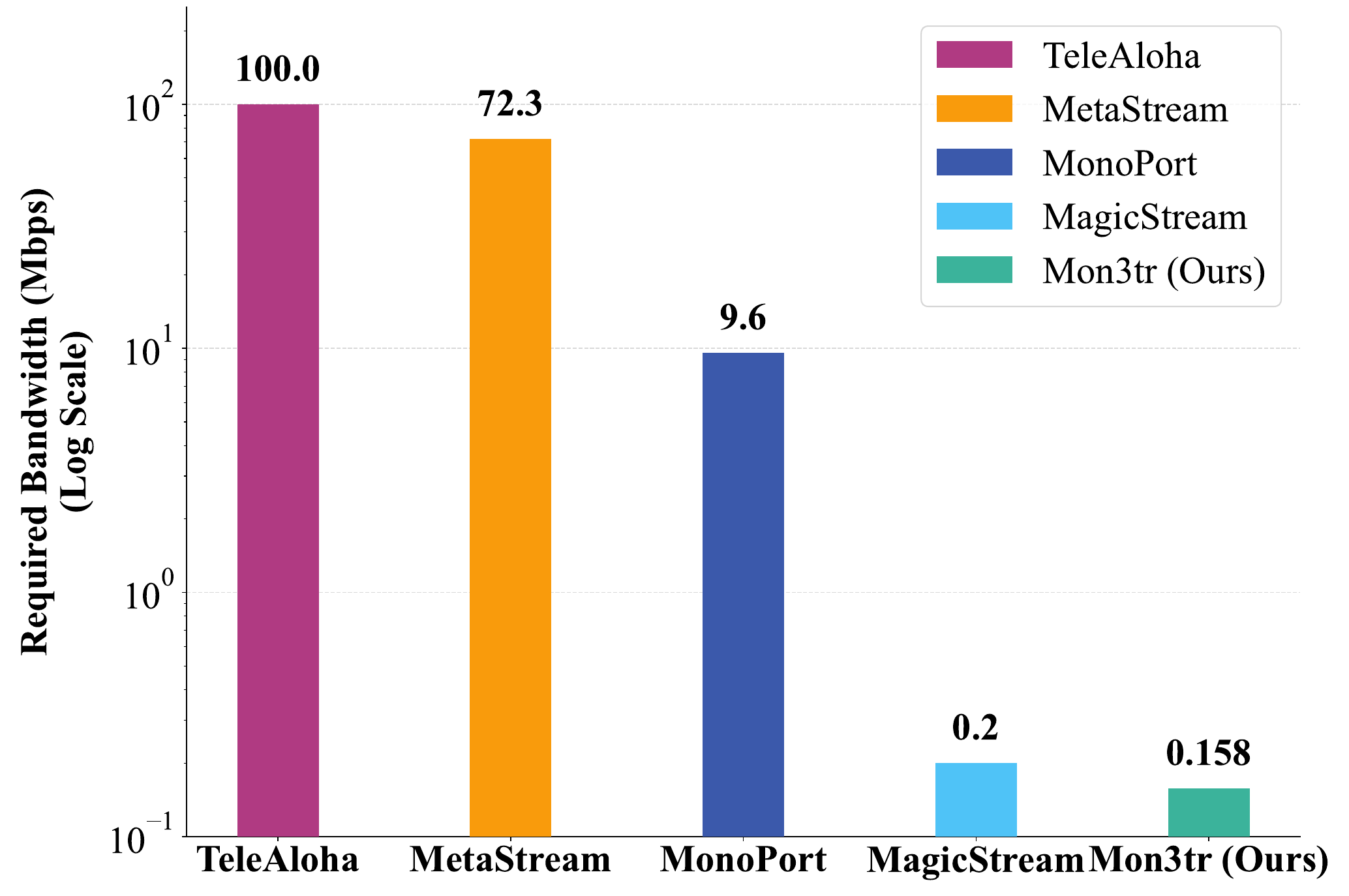}
    \caption{Comparison of bandwidth requirements between Mon3tr and baselines, where the unit for the values is Mbps.}
    \label{fig:bw}
\end{figure}

\textbf{Bandwidth Consumption.} 
The communication overhead is presented in Fig. \ref{fig:bw}, illustrating the impact of our ``computing for communication" design philosophy. By transmitting compact semantic parameters rather than raw 3D data, Mon3tr operates at an ultra-low bitrate of roughly 0.16 Mbps, which is even lower than MagicStream. This represents a magnitude reduction of over $1000\times$ compared to volumetric streaming systems like TeleAloha (100 Mbps) and MetaStream (72.3 Mbps), which struggle under network fluctuations. The result demonstrates that only delivering the semantic SPMM3 parameters is enough to maintain a natural and comfortable communication experience, which is favorable in bandwidth-constrained networks.

\begin{figure}[htbp]
    \centering    
    \includegraphics[width=0.95\columnwidth]{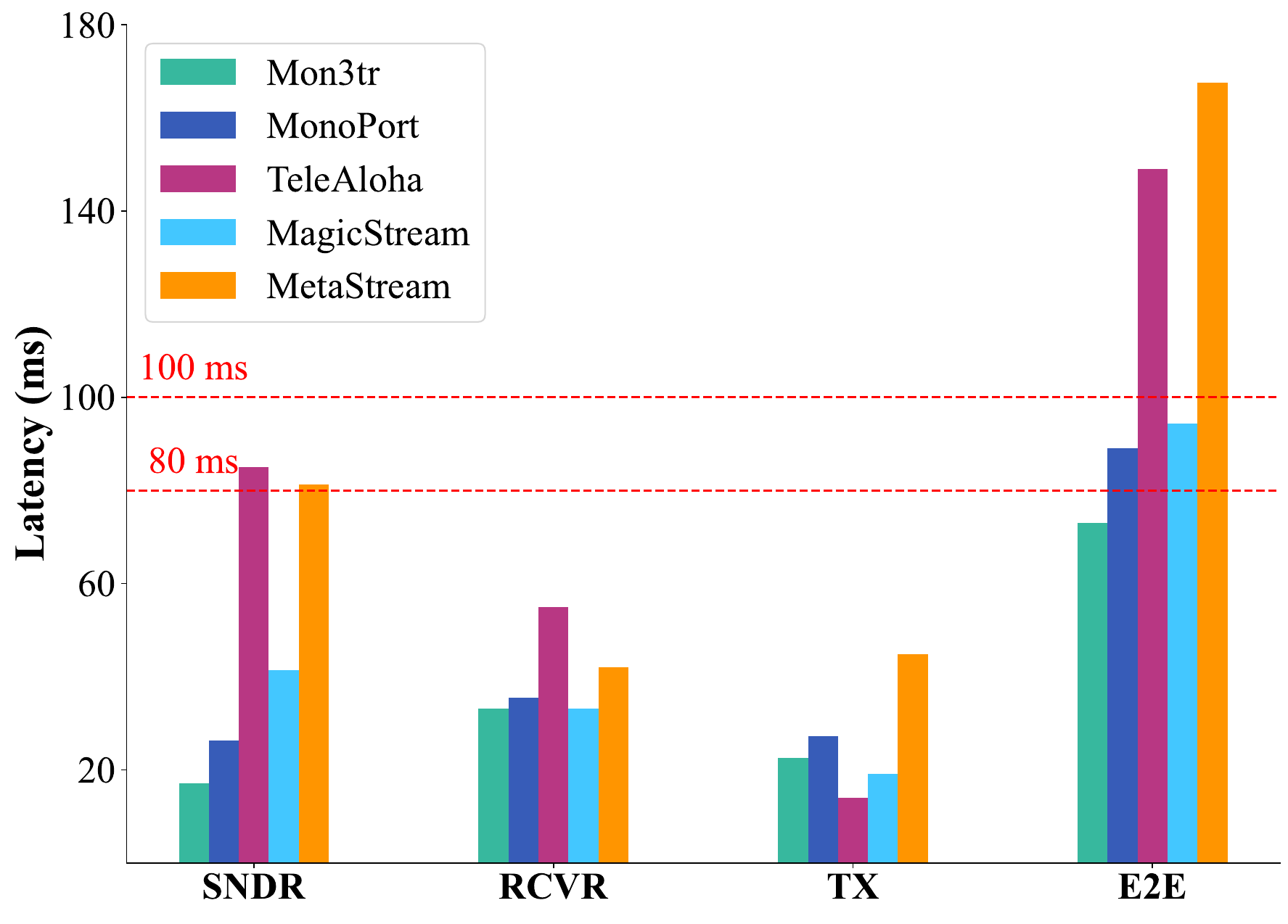}
    \caption{Comparison of end-to-end latency breakdown.}
    \label{fig:latency}
\end{figure}

\textbf{End-to-End Latency Evaluation.} 
To ensure natural interaction, the system latency must remain imperceptible. Fig. \ref{fig:latency} presents the end-to-end latency breakdown, showing that Mon3tr achieves a total delay of just 73.1 ms, comfortably below the 100 ms threshold required for seamless communication. This is less than half the latency of MetaStream (167.5 ms) and significantly faster than MonoPort (89.1 ms). Notably, our optimized sender-side parallel pipeline incurs only $\sim$ 21 ms for parameter extraction, vastly outperforming the 91 ms capture time of TeleAloha. This result confirms that utilizing offline amortization for heavy geometry processing allows the online phase to remain ultra-responsive, enabling true real-time responsiveness.

\section{Conclusion and Future Work}
\label{sec:conclusions}

In this paper, we present \textbf{Mon3tr}, a novel monocular 3D telepresence framework that solves the critical bottlenecks of bandwidth and hardware dependency in immersive communication. By pioneering an amortized computation paradigm, Mon3tr circumvents the prohibitive costs of traditional volumetric streaming. Our approach consists of a one-time offline phase to construct a high-fidelity personalized 3DGS-based avatar and a lightweight online phase where only compact SPMM3 motion parameters are transmitted from a single monocular RGB camera. This strategy drastically reduces bandwidth requirements by over $1000\times$ to less than $0.2$ Mbps.
On the receiver side, our system leverages lightweight deformation networks to animate the pre-built avatar, achieving a photorealistic rendering quality of over $28$ dB PSNR for novel poses at a real-time streaming rate of $\sim60$ FPS, with an end-to-end latency of under $80$ ms. By decoupling visual fidelity from per-frame computational and network load, Mon3tr makes high-quality holographic telepresence practical and accessible on consumer-grade hardware, paving the way for the next generation of remote interaction.

While Mon3tr achieves significant advancements, there remain opportunities to further enhance its capabilities. Currently, adapting user appearances, such as changes in clothing, requires new multi-view video recordings and an offline reconstruction process. This approach can be time-consuming and resource-intensive, limiting the diversity and flexibility of virtual avatar customization. By pre-training a generative model on extensive human datasets, one can build a 3D avatar with merely a single input image or a short clip of monocular video, further facilitating the amortization process. Furthermore, by utilizing a single image or a text prompt as input, these models should be able to directly edit the pre-built 3DGS avatar, enabling applications such as virtual try-on. This enhancement will facilitate a ``capture once, generate many appearances'' paradigm, significantly improving the potential for personalized user experiences in immersive telepresence and other applications within the metaverse.

\ifCLASSOPTIONcaptionsoff
  \newpage
\fi

\bibliographystyle{IEEEtran}
\bibliography{refs}

\end{document}